\documentclass[conference]{IEEEtran}
\IEEEoverridecommandlockouts
\usepackage{cite}
\usepackage{amsmath,amssymb,amsfonts}
\usepackage{algorithmic}
\usepackage{graphicx}
\usepackage{textcomp}
\usepackage{xcolor}
\usepackage[misc]{ifsym}
\usepackage{mathrsfs}
\usepackage{subfigure}
\usepackage{amsthm, amssymb}
\usepackage{stfloats}
\usepackage{float}
\usepackage{multirow}
\usepackage{color}
\usepackage{bm}
\usepackage{booktabs}
\usepackage{enumitem}
\usepackage{threeparttable}
\usepackage{amssymb}
\usepackage{amsmath}

\def\BibTeX{{\rm B\kern-.05em{\sc i\kern-.025em b}\kern-.08em
    T\kern-.1667em\lower.7ex\hbox{E}\kern-.125emX}}

\begin{document}

\title{Cooperative Multi-Agent Reinforcement Learning with Hypergraph Convolution}

% \title{Conference Paper Title*\\
% {\footnotesize \textsuperscript{*}Note: Sub-titles are not captured in Xplore and
% should not be used}
% \thanks{Identify applicable funding agency here. If none, delete this.}
% }

% \author{\IEEEauthorblockN{Yunpeng Bai}
% \IEEEauthorblockA{\textit{dept. name of organization (of Aff.)} \\
% \textit{Institude of authation chinese academy of sciences}\\
% baiyunpeng2020@ia.ac.cn}\\
% % City, Country \\
% % email address or ORCID}
% \and
% \IEEEauthorblockN{2\textsuperscript{nd} Given Name Surname}
% \IEEEauthorblockA{\textit{dept. name of organization (of Aff.)} \\
% \textit{name of organization (of Aff.)}\\
% City, Country \\
% email address or ORCID}
% \and
% \IEEEauthorblockN{3\textsuperscript{rd} Given Name Surname}
% \IEEEauthorblockA{\textit{dept. name of organization (of Aff.)} \\
% \textit{name of organization (of Aff.)}\\
% City, Country \\
% email address or ORCID}
% \and
% \IEEEauthorblockN{4\textsuperscript{th} Given Name Surname}
% \IEEEauthorblockA{\textit{dept. name of organization (of Aff.)} \\
% \textit{name of organization (of Aff.)}\\
% City, Country \\
% email address or ORCID}
% \and
% \IEEEauthorblockN{5\textsuperscript{th} Given Name Surname}
% \IEEEauthorblockA{\textit{dept. name of organization (of Aff.)} \\
% \textit{name of organization (of Aff.)}\\
% City, Country \\
% email address or ORCID}
% \and
% \IEEEauthorblockN{6\textsuperscript{th} Given Name Surname}
% \IEEEauthorblockA{\textit{dept. name of organization (of Aff.)} \\
% \textit{name of organization (of Aff.)}\\
% City, Country \\
% email address or ORCID}
% }

\author{Anonymous IJCNN 2022 submission\\
Paper ID: 531\\}

\author{\IEEEauthorblockN{Yunpeng Bai$^{*,1,4}$, Chen Gong$^{*,2,3,4}$, Bin Zhang$^{*,1,4}$, Guoliang Fan $^{1,4,}$ \Letter$\;$, Xinwen Hou$^{2,4}$, Yu Liu$^{2,4}$\thanks{* THESE AUTHORS CONTRIBUTED EQUALLY.}\thanks{THIS PROJECT WAS SUPPORTED BY NATIONAL DEFENCE FOUNDATION REINFORCEMENT FUND, NATIONAL DEFENSE BASIC SCIENTIFIC RESEARCH PROGRAM (JCKY2019203C029, JCKY2019-207B022) AND NATIONAL FOUNDATION FUND (Y2009-1B-02-353).}}
\IEEEauthorblockA{$^1$ Fusion Innovation Center, Institute of Automation, Chinese Academy of Sciences}
\IEEEauthorblockA{$^2$ Comprehensive information system research Center, Institute of Automation, Chinese Academy of Sciences}
\IEEEauthorblockA{$^3$ School of Computing and Information Systems, Singapore Management University}
$^4$ School of Artificial Intelligence, University of Chinese Academy of Sciences\\
% Beijing, China\\
\{baiyunpeng2020, gongchen2020, zhangbin2020, guoliang.fan, xinwen.hou, yu.liu\}@ia.ac.cn
}
\maketitle
\begin{abstract}
Recent years have witnessed the great success of multi-agent systems (MAS). 
Value decomposition, which decomposes joint action values into individual action values, has been an important work in MAS. %i.e. value decomposition.
However, many value decomposition methods ignore the coordination among different agents, leading to the notorious ``lazy agents'' problem.
To enhance the coordination in MAS, this paper proposes \emph{HyperGraph CoNvolution MIX}~(HGCN-MIX), a method that incorporates hypergraph convolution with value decomposition. HGCN-MIX models agents as well as their relationships as a hypergraph, where agents are nodes and hyperedges among nodes indicate that the corresponding agents can coordinate to achieve larger rewards. Then, it trains a hypergraph that can capture the collaborative relationships among agents. Leveraging the learned hypergraph to consider how other agents' observations and actions affect their decisions, the agents in a MAS can better coordinate.
We evaluate HGCN-MIX in the StarCraft II multi-agent challenge benchmark.
The experimental results demonstrate that HGCN-MIX can train joint policies that outperform or achieve a similar level of performance as the current state-of-the-art techniques. We also observe that HGCN-MIX has an even more significant improvement of performance in the scenarios with a large amount of agents. Besides, we conduct additional analysis to emphasize that when the hypergraph learns more relationships, HGCN-MIX can train stronger joint policies. 
\end{abstract}


\section{Introduction}
\label{intro}
Deep Reinforcement Learning (DRL), which uses deep neural networks (DNN) to learn an optimal policy, has increasingly gained attention from researchers and achieved successful applications in various fields, including video games~\cite{mnih2015human, silver2016mastering, gong2020stable,DBLP:conf/icmcs/GongHBHFL21}, robotics~\cite{levine2016end}, etc.

In reality, many tasks require multiple agents, e.g. autonomous vehicles~\cite{rahmati2019uw}, resource allocation problems~\cite{cui2019multi}, etc. The single-agent RL algorithms ignore the coordination between agents, such as the independent $Q$ learning (IQL)~\cite{tampuu2017multiagent}, which demonstrates poor performance in multi-agent systems (MAS). Therefore, enhancing coordination in MAS becomes an essential direction to improve the performance of multi-agent reinforcement learning (MARL).
MARL algorithms train a joint policy~\cite{zhang2021fop} to maximize the cumulative rewards in a Markov Decision Process (MDP). Due to the non-stationarity problem~\cite{papoudakis2019dealing}, it is hard to explain how the environments feedback to the agents a reward based on the agents' actions. 
Value decomposition~\cite{Sunehag2018ValueDecompositionNF} is proposed based on the centralized training with decentralized execution (CTDE) paradigm~\cite{lowe2017multi}, to study that how individual action values (introduced in Section~\ref{decpomdp}) affect joint action values.
In CTDE, during the training phase, each agent acquires global information, but is not dependent on the global information during the execution phase~\cite{lowe2017multi}. It is noticed that the global information includes individual observation, actions, from other agents. 
\begin{figure}
\centering
\includegraphics[width=3.5 in]{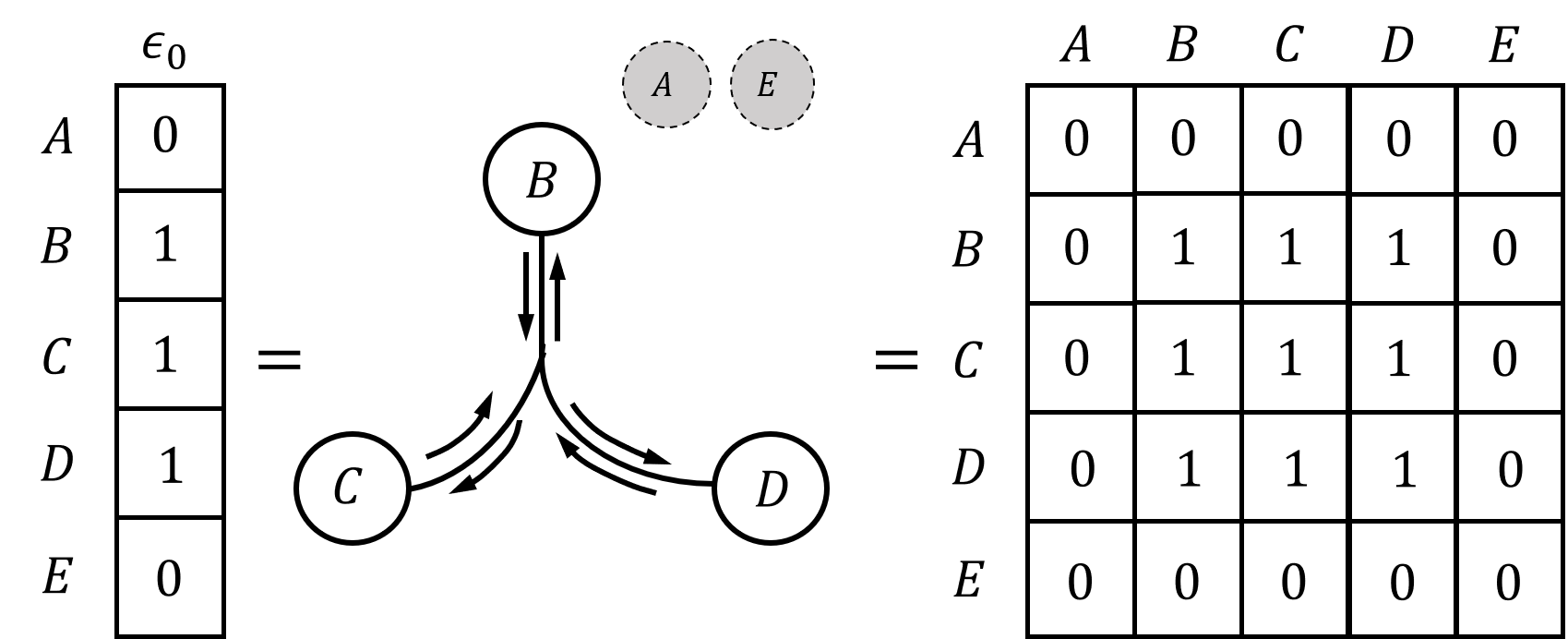}
\caption{As we shown in the mid figure, the hyperedge $\epsilon_0$ in the left part connects nodes $B, C, D$ and disconnects $A$ and $E$. The mid figure also presents that how neighborhood information on the hyperedge $\epsilon_0$ flows. And the right part shows a normal graph that equals to hyperedge $\epsilon_0$.}
\label{hyperedgeeqgraph}
\end{figure}
Value-Decomposition Networks (VDN) ~\cite{Sunehag2018ValueDecompositionNF} and QMIX~\cite{rashid2018qmix} , which are proposed on top of value decomposition (explained in Section~\ref{value_decomposition}), do not consider cooperation between multi-agents. Insufficient cooperation causes the notorious issue, i.e., ``lazy agents''~\cite{rashid2018qmix} who take non-optimal actions due to the wrong credit assignment. For example, 6 of 10 agents have learned optimal or sub-optimal policies, but the rest four agents learn nothing.

To alleviate this ``lazy agents'' problem, several studies (CommNet~\cite{sukhbaatar2016learning}, DGN~\cite{jiang2018graph})
% BicNet~\cite{peng2017multiagent} 
utilize the observations or embedding of recurrent neural network (RNN) transmitted by neighbor agents to enhance communication. Through better communication, agents obtain more information about other agents, especially other agents' observations and actions; thus, DGN~\cite{jiang2018graph} and CommNet~\cite{sukhbaatar2016learning} can significantly improve prior state-of-the-art works.
For each agent, collecting the information from other agents, i.e., observations, actions, is an important part to enhance the coordination of MAS. 
Graph Convolutional Network (GCN) achieves the integration by aggregating and propagating neighborhood information for those irregular or non-euclidean nature of graph-structured data~\cite{kipf2016semi}. Through GCN, we learn the embedding of each node or the whole graph based on the geometric structure. 
As shown in Figure~\ref{agentmodule}, in most value decomposition methods, such as QMIX, QPLEX~\cite{wang2020qplex}, individual observations and actions are transmitted to individual action values through RNN.
We take individual action values as the embedding of each agents' observations and actions. Treating each agent as each node in a graph and performing GCN, the embedding of agents' observations and actions is attached with neighborhood information.
However, it is difficult to represent the complex connections between a large of agents with a simple graph for these two reasons: (1) there is no prior knowledge for connections between agents to build the graph, and (2) connections between agents have a dynamic change due to the uncertainty of environments, observations, and global states. 
Hypergraphs~\cite{feng2019hypergraph, bai2021hypergraph} can overcome these shortcomings.
In hypergraphs, the value of hyperedge denotes the degree of connection between nodes. Specifically, the large weight represents a strong connection and vice versa. %Besides, hypergraphs modify the connections by adding different hyperedges.
As presented in Figure~\ref{hyperedgeeqgraph}, we list a simple example that equals a simple hyperedge to a regular graph.
As for the uncertainty of environments, we use neural networks that take dynamic individual observations as input to learn the connections in our hypergraphs. Thus, hypergraphs are automatically adjusted by different observations.

As the analysis mentioned above, our contributions are summarized as follows, we proposed an end-to-end method termed HGCN-MIX, which combines value decomposition and hypergraph convolution (HGCN), aiming at enhancing the coordination between agents via HGCN. In HGCN-MIX, we build the hypergraphs generated via agents' individual observations without any prior knowledge. Besides, compared with the traditional fixed and binary hypergraph, in HGCN-MIX the hypergraphs are non-binary and dynamic adjust with the change of agent's observations.
We perform our algorithms on 12 different scenarios in SMAC~\cite{samvelyan2019starcraft}. The experimental results show that our method outperforms or achieves a similar level of performance as the state-of-art MARL approaches in most scenarios. Especially in $MMM2$, and $27m\_vs\_30m$, HGCN-MIX obtains a higher 20\% winning rate than QMIX. Besides, we conduct an ablation study to research the influence of using hypergraphs with different number of hyperedges. Ablation emphasizes that connections, represented as hyperedges, play an important role in training a strong joint policy.

\section{Related Works}
This section discusses several works in MARL and value decomposition methods briefly, and then introduces the developments of GCN and HGCN.
\subsection{Multi-Agent Reinforcement Learning (MARL)}
Various researches contribute to overcoming the multi-agent cooperation challenge. 
Indepent Q leanring (IQL)~\cite{tampuu2017multiagent} extends the architecture of Deep Q-Learning Network (DQN)~\cite{mnih2015human} to MAS.
In mean field MARL, considering that each agent’s effect on the overall population can become infinitesimal~\cite{yang2018mean}.
Although mean filed MARL has demonstrated excellent performance in mean field games (MFGs), MFGs require homogeneous agents~\cite{lasry2007mean}. 
Recently, based on the centralized training
with decentralized execution (CTDE) paradigm, some remarkable approaches such as value function decomposition~\cite{Sunehag2018ValueDecompositionNF}, and multi-agent policy gradient methods~\cite{yu2021surprising} have been proposed.
Compared with value-based methods, policy gradient methods can be easily applied to the environment with continuous actions space.
In Multi-agent deep deterministic policy gradient (MADDPG)~\cite{lowe2017multi}, the critic network takes overall individual observations and actions as inputs.
% Multi-Actor-Attention-Critic (MAAC)~\cite{iqbal2019actor} replaces DDPG~\cite{lillicrap2015continuous} in MADDPG as Soft Actor-Critic (SAC)~\cite{haarnoja2018soft}, and uses attention mechanism to learn the centralized critic network.
As mentioned in~\cite{papoudakis2021benchmarking}, although policy gradient (PG) methods significantly outperform the QMIX in some continuous environments (multi-agent particle world environment (MPE)~\cite{lowe2017multi}), value-based methods achieve more sample-efficient than PG approaches in some discrete environments (SMAC).
\vspace{-0.05 in}
\subsection{Value Decomposition Methods for MARL}
\label{vd_realted}
Value decomposition methods decompose the joint state-action value function into a single global reward.
On the basis of CTDE, plenty of MARL algorithms are proposed, e.g., Value-Decomposition Networks (VDN)~\cite{Sunehag2018ValueDecompositionNF}, QMIX ~\cite{rashid2018qmix}, QPLEX~\cite{wang2020qplex}. 
VDN denotes the joint action value as the sum of individual action values.
QMIX approximates the joint state-action value function monotonously.
MMD-MIX~\cite{xu2021mmd} applies MMD-DQN~\cite{tang2020distributional} to collect randomness information in situations. Qtran~\cite{son2019qtran} relaxes the constraints of joint action-value functions, but has a poor performance in some complex tasks. 
The most related to our work is GraphMIX~\cite{naderializadeh2020graph}, which combines graph neural networks (GNNs)~\cite{scarselli2008graph} and value decomposition. However, GraphMIX only succeeds in several particular scenarios (i.e., $6h\_vs\_8z$).
\vspace{-0.05 in}
\subsection{Graph Convolutional Network (GCN) and Hypergraph Convolutional Network (HGCN)} 
% For those irregular or non-Euclidean nature of the graph-structured data, GCN 
GCN is proposed in~\cite{kipf2016semi} for classification tasks. GCN aims to capture neighborhood information for the irregular or non-euclidean nature of graph-structured data.
GraphSAGE leverages the node feature information~\cite{hamilton2017inductive}.
Only sampled neighbor nodes features are used for leveraging through aggregating functions.
ChebNET~\cite{defferrard2016convolutional} performs Fourier transform and converts spatial graph data into Fourier domain.
Although GCN has been widely used in many areas, such as event detection~\cite{nguyen2018graph} and  knowledge graph~\cite{wang2018cross}, we select HGCN to aggregate instead of GCN.
Replacing graphs with hypergraphs, Bai et al. deviate the two forms of convolution in HGCN~\cite{bai2021hypergraph}.
\cite{jiang2019dynamic} performs k-means clustering method to construct hypergraphs.
The hypergraph attention mechanism has also been proposed in~\cite{bai2021hypergraph}. Hypergraphs in the hypergraph attention mechanism are also dynamic and non-binary. 
Reasons about our selection are elaborated in Section~\ref{reasonsforhg}.

\section{Background}
In this section, we briefly reviews about the basis of Dec-POMDP, value decomposition, and the definitions of hypergraph and hypergraph convolution.
\vspace{-0.05 in}
\subsection{Decentralized partially observable Markov
decision process (Dec-POMDP)}
\label{decpomdp} 
Dec-POMDP~\cite{oliehoek2016concise} describes the fully cooperative multi-agent tasks.
A Dec-POMDP can be defined by a tuple $G=\langle \mathcal{S}, \mathcal{U}, \mathcal{P}, \mathcal{Z}, r, \mathcal{O}, n, \gamma \rangle$. 
$s\in\mathcal{S}$ denotes the global state of the environment. Each agent $a \in \mathcal{A}:= \{1,\dots, n\}$ takes an individual action $u_a \in \mathcal{U}$ at each timestep. $\mathcal{U}$ is the action set. All the individual actions in the same timestep form a joint action $\mathbf{u} \in \boldsymbol{\mathcal{U}}\equiv \mathcal{U}^n$. $\mathcal{P}(s'|s, \mathbf{u}): \mathcal{S}\times \boldsymbol{\mathcal{U}} \to \Delta(\mathcal{S})$ denotes the transition probability for any joint actions of both agents, where $\Delta(\cdot)$ is the simplex.
All the agents share the same reward function $r(s,\mathbf{u}): \mathcal{S} \times \boldsymbol{\mathcal{U}} \xrightarrow{}\mathbb{R}$.

Each agent gets its own local individual partial observation $z\in\mathcal{Z}$ according to the observation function $\mathcal{O}(s,a): \mathcal{S} \times \mathcal{A} \xrightarrow{}\mathcal{Z}$. The previous of actions and observations for a agent makes up its own action-observation history $\tau_a \in T \equiv (\mathcal{Z} \times \mathcal{U})$. Policy of agent $a$ is represented as: $\pi_a(u_a|\tau_a): T\times\mathcal{U}\xrightarrow{}\left[0,1\right]$. The cumulative discounted reward in timestep $t$ is defined as $R^t = \sum_{l=t}^{\infty}\gamma^{l-t} r_{l}$~\cite{sutton2018reinforcement}, and we aim to maximize the expected reward $\mathbb{E}_{\tau \sim \pi}[R^0|s_0]$, where $\tau$ is a trajectory and $\gamma \in [0,1)$ is the discount factor.
\vspace{-0.1 in}
\subsection{Value decomposition}
\label{value_decomposition}
In Dec-PODMP, agents share the same joint reward, leading to the design of a unique reward function to train independent agents difficultly. As we have discussed in Section~\ref{intro}, directly applying DQN for each agent without any coordination may lead to ``lazy agents"~\cite{rashid2018qmix}.

Based on the CTDE paradigm, value decomposition methods learn the decomposition of the joint value function. 
The core of CTDE is that a centralized critic (named ``mixing network'' in Figure~\ref{structfig}) guides the agents during training. This critic only exists in the training phase and has the ability to access global states, individual observations, individual action values, and actions. Besides, during the training phase, each agent can acquire neighborhood information, including individual observation, actions from other agents.
During the execution phase, the centralized critic is deleted from the structure, and agents are not allowed to access neighborhood information. 
There are several meaningful works proposed based on value decomposition, such as VDN, and QMIX.
One of the most important equation in value decomposition is Individual-Global-Max (IGM) equation~\cite{son2019qtran}:
\begin{equation}
\arg \max _{\boldsymbol{u}} Q_{\mathrm{tot}}(\boldsymbol{\tau}, \boldsymbol{u})=\left(\begin{array}{c}
\arg \max _{u_{1}} Q_{1}\left(\tau_{1}, u_{1}\right) \\
\vdots \\
\arg \max _{u_{n}} Q_{n}\left(\tau_{n}, u_{n}\right)
\end{array}\right).
\label{IGMEQ}
\end{equation}
IGM assumes that the optimality of each agent is consistent with the global optimality. Almost all algorithms based on value decomposition satisfy
the IGM condition and finally achieve convergence.
QMIX approximates the joint
state-action value function monotonously:
\begin{equation}
    \frac{\partial Q_{tot}(\tau, u)}{\partial Q_a(\tau_a, u_a)} \geq 0, \forall a \in {1,\dots,n},
\end{equation}
where $\tau_a$ denotes action-observation history of agent $a$, and $u_a$ denotes individual action of agent $a$.
The joint value functions in Qtran is formulated as:
\begin{equation}
Q_{tot}(s, \textbf{u}) = \phi(s, Q_0, (\tau_0, u_0), \dots, Q_{n}, (\tau_n, u_n)),
\end{equation} 
where $\phi$ varies in each method. In some complex tasks, Qtran has a poor performance. These methods (e.g., VDN, QMIX, etc.) pay more attention on decomposing the joint value function, but ignore the coordination between agents.
\subsection{Definition of Hypergraph}
\label{hyperdef} 
A hypergraph~\cite{berge1984hypergraphs} $\mathcal{G}$ is consisted of a vertex set $\mathbf{V}= \{v_1, v_2,\dots,v_N\}$ ($N$ represents the number of vertices) and the edge set $\mathbf{E}$ with $M$ hyperedges. The adjacency matrix of the hypergraph $H\subseteq \mathbb{R} ^{N\times M}$ indicates the connection between a vertice and hyperedge. For a vertex $v_i\in\mathbf{V}$ and a hyperedge $\epsilon\in\mathbf{E}$, when $v_i$ is connected by $\epsilon$, $H_{i\epsilon} = 1$ exists, otherwise we have $H_{i\epsilon} = 0$. Besides, each hyperedge $\epsilon$ has a non-negative weight $w_{\epsilon\epsilon}$. All the $w_{\epsilon\epsilon}$ is form a diagonal matrix $\mathbf{W} \in \mathbb{R}^{M\times M}$. 
In a hypergraph, the vertex degree $\textbf{D}\in \mathbb{R}^{N\times N}$ and the hyperedge degree $\textbf{B} \in \mathbb{R}^{M\times M}$ are defined as 
% \begin{equation*}
$D_{ii} = \sum_{\epsilon=1}^{M} w_{\epsilon \epsilon}H_{i\epsilon},\; B_{\epsilon\epsilon} = \sum_{i=1}^{N}H_{i\epsilon}$.
% \end{equation*}
It is noticed that $\textbf{D}$ and $\textbf{B}$ are diagonal matrices.

In an indirect graph, the degree of an edge is 2, presenting that each edge only connects 2 vertices. However, in a hypergraph, an edge may connect more than 2 vertices. Thus, the property of the hypergraph allows more information about the nodes included in the hyperedge. Furthermore, it is easy to incorporate prior knowledge of connections between agents by selecting various hyperedges.

\subsection{Spectral convolution on hypergraph}
Given a hypergraph $\mathcal{G} = (\bm{V}, \bm{E}, \bm{\Delta})$ with $N$ vertices and $M$ hyperedges, $\bm{\Delta} \in \mathbb{R}^{N\times N}$ is the hypergraph Laplacian positive semi-definite matrix~\cite{feng2019hypergraph}. Then, $\bm{\Delta}$ is factorized as $\bm{\Delta}= \bm{\Phi}\bm{\Lambda}\bm{\Phi}^T$, where $\bm{\Phi}=\text{diag}(\phi_1, \dots, \phi_n)$ is an orthonormal eigen vectors and $\bm{\Lambda}=\text{diag}(\lambda_1, \dots, \lambda_n)$ is an diagonal matrix whose diagonal elements are the corresponding eigenvalues. The Fourier transform~\cite{bracewell1986fourier, kipf2016semi} for a signal $\bm{x} = (x_1, \dots, x_n)$ is defined as $\hat{\bm{x}}=\bm{\Phi}^T\bm{x}$. Then, the spectral convolution of signal $\bm{x}$ and filiter $\bm{g}$ is defined as:
\begin{equation}
\bm{g} * \bm{x} = \bm{\Phi}{g(\bm{\Lambda})}\bm{\Phi}^T\bm{x},
\end{equation}
where $g(\bm{\Lambda})=\text{diag}(\bm{g}(\lambda_1),\dots, \bm(\lambda_n)$ is a function of the Fourier coefficients.
However, the computing of our method incurs $\mathcal{O}(n^2)$ computational complexity. To solve this problem, Feng et al.  \cite{feng2019hypergraph, defferrard2016convolutional, kipf2016semi} derive spectral convolution as 
$$\bm{g} * \bm{x} \approx \theta_0\bm{x} - \theta_1\textbf{D}^{-\frac{1}{2}}\textbf{H}\textbf{W}\textbf{B}^{-1}\textbf{H}^T\textbf{D}^{-\frac{1}{2}}\bm{x},$$
where $\theta_0$ and $\theta_1$ are the parameters of filtering over all nodes. Further, with the utilization of a single parameter $\theta$ is defined as:
\begin{equation*}
\left\{
\begin{array}{l}
\theta_1 = -\frac{1}{2}\theta\\
\theta_0 = \frac{1}{2}\theta\textbf{D}^{-\frac{1}{2}}\textbf{H}\textbf{B}^{-1}\textbf{H}^T\textbf{D}^{-\frac{1}{2}}
\end{array}.
\right.
\end{equation*}
Then, the hypergraph convolution (HGCN) is derived as:
\begin{equation}
\bm{x}^{(l+1)} = \textbf{D}^{-\frac{1}{2}}\textbf{H}\textbf{W}\textbf{B}^{-1}\textbf{H}^T\textbf{D}^{-\frac{1}{2}}\bm{x}^{(l)}\textbf{P}, 
\label{HGCNEQ}
\end{equation} 
where $\textbf{P}$ denotes the weight matrix between the $(l)-$th and $(l+1)-$th layer~\cite{bai2021hypergraph}.
For each node in HGCN, more information can be accessed between those nodes connected by a common hyperedge. The greater the weights, the more information about observations, histories and actions is provided.

\section{Method}
In this section, we first introduce the reasons that we select HGCN instead of GCN or hypergraph attention. Secondly, we demonstrate the difficulties of traditional hypergraphs and our method to construct hypergraphs. Then, we introduce our model, HGCN-MIX, an end-to-end method base on CTDE paradigm.
Finally, we have a brief discussion about the loss function that follows the TD-error~\cite{sutton2018reinforcement}.
\begin{figure}[!t]
\centering
\includegraphics[width = 3.1 in]{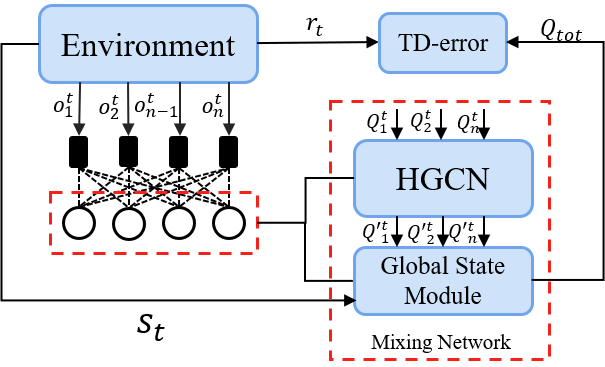}
\caption{The architecture of HGCN-MIX. At timestep $t$, agents first get their individual observations $\textbf{o}^t$ and output the individual action values $\textbf{Q}$. Following Eq.~(\ref{genH}), the hypergraph are generated with $\textbf{o}^t$ for HGCN. Then, HGCN takes individual action values $\textbf{Q}$ as the input and output the transformed action values $\textbf{Q'}$. The ``global state module" takes the global state $s_t$ and $\textbf{Q'}$, and outputs the $Q_{tot}$. The $Q_{tot}$ and $r_t$ are used to calculate the ``TD-error". }
\label{structfig}
% \vspace{-0.05in}
\end{figure}

\subsection{Why hypergraph convolution}
\label{reasonsforhg}
\subsubsection{Why hypergraph}
\label{whynotgraph}
As discussed in Section~\ref{intro}, GCN has the ability to aggregreate neighbourhood information, but compared with the graph, 
there are two main reasons to select the hypergraph: 
(1) there are no prior knowledge about the connectivity between nodes (which denote agents) to build a regular graph. (2) relationships between agents have a dynamic change due to
the dynamic environments, observations, and global states. Therefore, we need multiple graphs to learn the different relationships between agents. Due to a regular graph is equal to a hyperedge, as we shown in Figure~\ref{hyperedgeeqgraph}, it is easily for us to build the multiple hyperedges representing multiple regular graphs.

\subsubsection{Why not attention}
\label{whynotattention}
It is difficult to learn the connections between agents for value decomposition methods in hypergraph attention mechanism because of two disadvantages: (1) hypergraph attention mechanism takes the pairwise similarity as the weights of connections in a hypergraph. (2) hypergraph attention takes the encoding of nodes as input.
Nodes with different pairwise similarity get low weights even if the connections between these nodes are strong. For example, the two cikissis in $2c\_vs\_64zg$ may take opposite actions to coordinate. The encoding of their observations and actions may be huge different but connections between them are quite strong.
We aim at finding more different connections between agents, so pairwise similarity is not suitable here. 
In value-decomposition methods, individual action values are essential factors for selecting actions. Besides, the learning of embedding takes additional computational resource.

% while HGCN-MIX takes individual action values.
% There are several reasons that selects hypergraph convolution for aggregation in HGCN-MIX rather than hypergraph attention mechanism. On the one hand, hypergraph attention takes the pairwise similarity as the weights of connections in a hypergraph.
% Similar pairwise vertices get higher weights than those .
% We aim at finding more different connections between agents, so pairwise similarity is not suitable here. On the other hand, hypergraph attention takes the encoding of nodes as input, while HGCN-MIX takes individual action values. In value-decomposition methods, individual action values are essential factors for selecting actions.
% What's more, hypergraph attention updates weights in hypergraphs, while HGCN-MIX updates the neural network.
% The neural network in HGCN-MIX takes dynamic observations and actions as input and generates the hypergraphs.
% In this way, hypergraphs can hardly take current information hidden in agents' individual observations and actions.
\subsection{Difficulties of traditional hypergraphs}
In Section~\ref{hyperdef}, we elaborate the definition of hypergraphs.  Traditional hypergraphs usually have three drawbacks: (1) prior knowledge is required to build hypergraph, but the prior knowledge is usually hard to construct; and (2) the hypergraph is fixed when it is established, which leads to fixed connections between agents; and
3) the hypegraph only include two value (i.e., 1 or 0), indicating whether the node is connected by the hyperedge.
% In the binary value hypergraphs, the hypergraph only reflects two kinds of relationships, which means the hypergraph will only show whether the agent will be connected by other agents. 
Based on the analysis mentioned above, we rethink that agents need different levels of collaboration in different situations, e.g., in $MMM2$ agents need a high-level collaboration, but only low-level collaboration is needed for agent in $2s3z$. Surely, the traditional hypergraph can not solve these problems we mentioned. 

Instead of building a fixed binary hypergraph beyond prior knowledge, we generate a dynamic hypergraph via neural networks which adjusts automatically according to the agents' individual observations. We use real-value hypergraph to describe the connections between agents, where different values of hyperedges represent different levels of collaboration. 

\subsection{Building hypergraph}
\label{buildhyper}
To describe the degrees of collaboration appropriately, we replace the original binary hypergraph with a real-value hypergraph. As mentioned in~\cite{bai2021hypergraph}, non-binary hypergraphs also follow Eq.~(\ref{HGCNEQ}). 

In HGCN-MIX, a linear function is utilized, which takes individual observation $z_i\in\mathcal{Z}, i \in\{1,\dots,n\}$ as input, and generates a real values vector $\mathrm{A_i} \in \mathbb{R}^m$, where $m$ denotes the number of hyperedges. Then, under the ReLU operation, the negative weights are modified to 0. On the one hand, negative weights break the IGM equation (Eq.~(\ref{IGMEQ})). On the other hand, negative weights make the hypergraph irreversible.
Each value in $\mathrm{A_i}$, denoted as $\omega_{i,j}, j \in \{1,\dots,m\}$, presents the weights for hyperedge $\epsilon_j$.
The entirety of $\omega_{i,j}$ consists of the first part $H_1$ of our hypergraph $H$:
\begin{equation}
    H_1 = \text{ReLU}(\text{Linear}(\mathcal{Z})).
    \label{genH}
\end{equation}
The above equation lead agents to collaborate with others in different observations.

Inspired by GCN, we add an $n$-dimensional identity matrix, which is denoted as $H_2 \subseteq \mathbb{R}^{n\times n}$, and $n$ is the number of agents. $H_2$ makes agents concentrate on themselves and maintain track of their own information. It is noticed that compared with $H_1$, a fixed value is impossibly fitness for all the environments, e.g. in $3m$, $1$ is too large but too small for $MMM2$. Therefore, we multiply $H_2$ by the average of $H_1$ to make the  $H_2$ having the same orders as the $H_1$’s. Thus, we formulate $H_2$ as:
\begin{equation*}
    H_2 = \mathbb{I}_n \times \text{avg}(H_1).
\end{equation*}
In some environments with weak cooperation or few account of agents, $H_2$ matrix paves the path for keeping their own information.

We denote the average of $H_1$ as $\mu$.
Then, joining the two $H_1$ and $H_2$ matrices together, we reformulate the final hypergraph $H \subseteq \mathbb{R}^{n\times (m+n)}$ as:
\begin{equation}
H=
\underbrace{\left[\begin{array}{ccc}
\omega_{1,1},&\dots,&\omega_{1,m}\\
\omega_{2,1},&\dots,&\omega_{2,m}\\
\vdots,&\cdots,&\vdots\\
\omega_{n,1},&\dots,&\omega_{n,m}
\end{array}\right]}_{H_1:\;Learning} \underbrace{\left[\begin{array}{ccc}
\mu,&\cdots,&0\\
0, &\cdots,&0\\
\vdots,&\cdots,&\vdots\\
0,&\cdots,&\mu
\end{array}\right]}_{H_2:\;One-hot},
\label{HYPERGRAPH}
\end{equation}
where $\omega_{i,k}, i\in[1, n], \text{and} \  k \in [1, m]$ is the weight of learned connection between agents, and $n$ denotes the number of agents and $m$ denotes the number of hyperedges, $\mu$ is the average value of $H_1$.

\subsection{Mixing network}
\begin{figure}[!t]
\centering
\subfigure[Agent Module]{
\includegraphics[width=1.4 in]{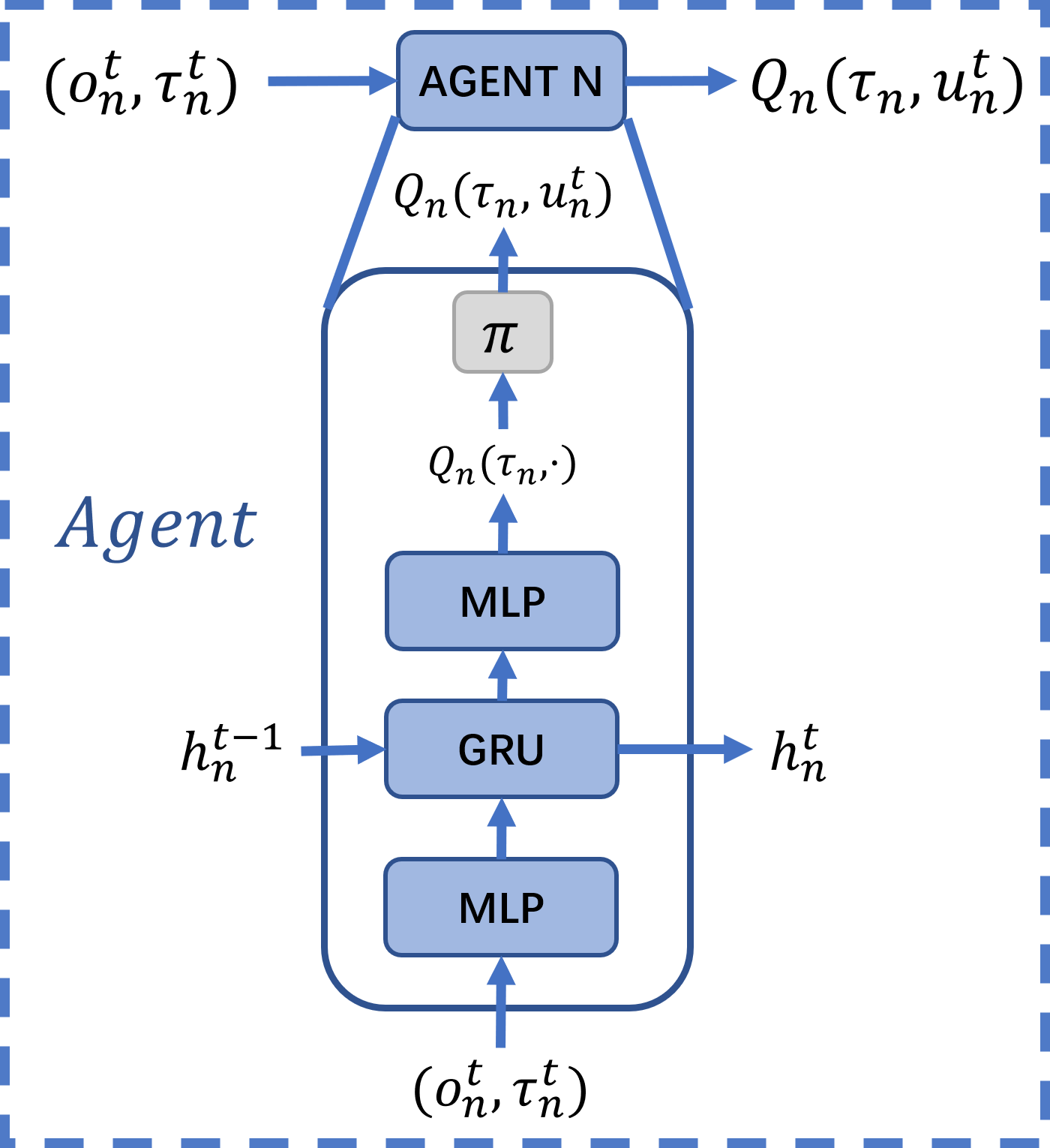}
\label{agentmodule}
}
\subfigure[Global State Module]{
\includegraphics[width=1.69 in]{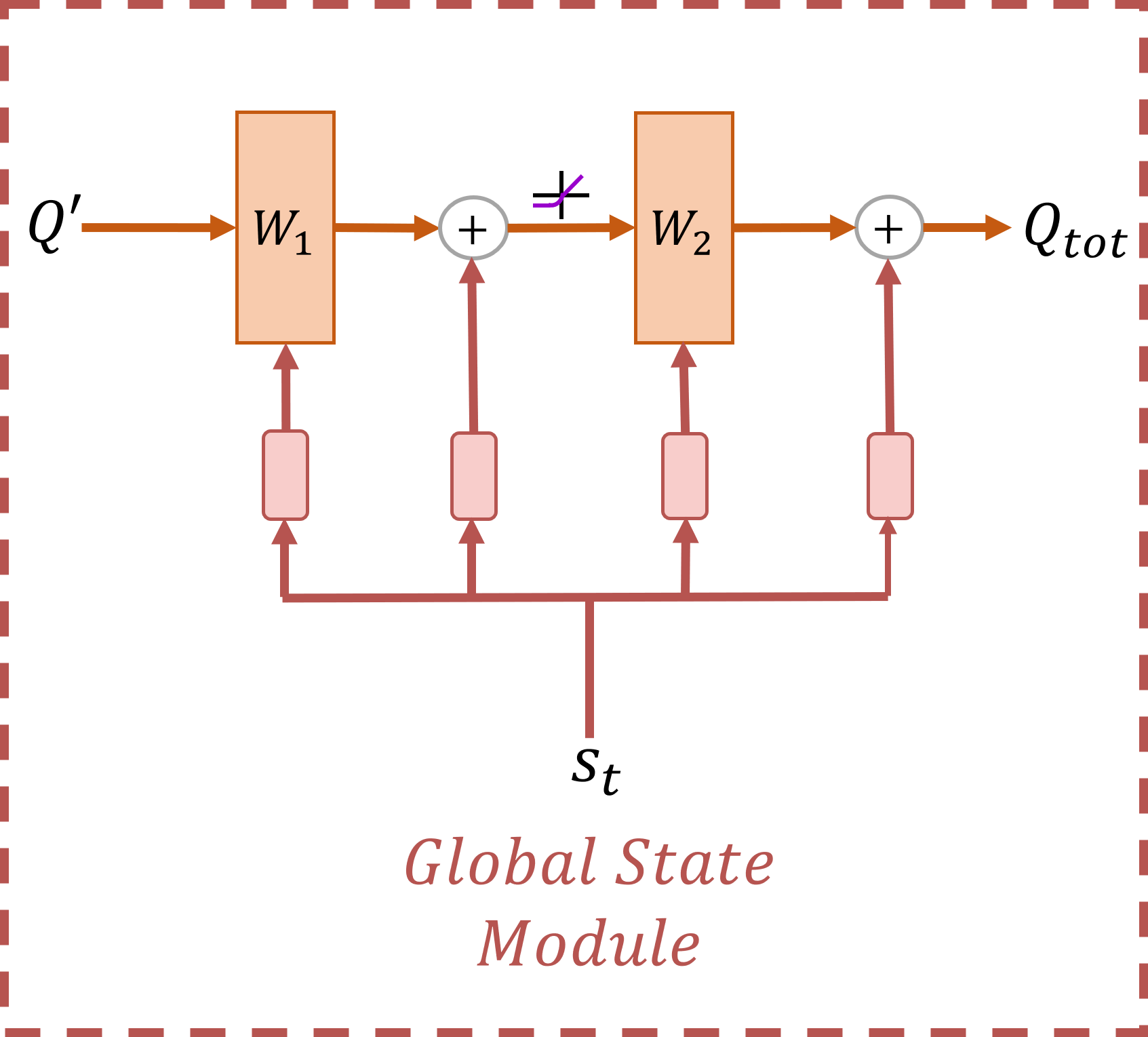}
\label{statemodule}
}
\caption{Figure~\ref{agentmodule} shows the structure of the agent module.
At timestep $t$, agent $n$ take individual observation $o_n^t$ and history $\tau_n^t$ as input, and output \textbf{Q}$(\tau_n, \cdot)$ the values of legal actions allowed in the environment. Then, individual $Q$ value $Q_n(\tau_n, u_n^t)$ is generated through the $\epsilon$-greedy.
Figure~\ref{statemodule} shows the process to add the global state information. At timestep $t$, by applying four different MLPs and activation functions, global state $s_t$ is transformed into four parts. Following Eq.~(\ref{stateeq}), two of four parts have a vector inner product with transformed action values \textbf{$Q'$}. The rest two parts are added to \textbf{$Q'$}. Then, the joint action value $Q_{tot}$ is generated.
}
\end{figure}

The structure of HGCN-MIX is presented in Figure~\ref{structfig}. 
As shown in Figure~\ref{agentmodule}, each agent applies a DRQN ~\cite{hausknecht2015deep} to learn individual action value $Q_a \in \bm{Q}$, where $a\in\{1,\dots,n\}$. Then, the hypergraph convolution performs on $\bm{Q}$. In convolution, we set $\textbf{P}$ to $\mathbb{I}_n$, and $\mathbf{W}$ to the learned parameters. To keep Eq.~(\ref{IGMEQ}), we use the absolute value of $\bm{W}$ for HCGN,
\begin{equation*}
    \bm{Q}' = \text{HGCN}(\text{HGCN}(\bm{Q}, \text{abs}(\mathbf{W}_1)), \text{abs}(\mathbf{W}_2)),
\end{equation*}
where $\mathbf{W}_i, i \in \mathbb{R}$ represents the weight matrix in the $i$-th layer and $\bm{Q}' \in \mathbb {R}^{n}$ denotes the modified action values.
Each action value $Q_a$ appends the information contained in all the $Q_a$ by our method, so that agents can strengthen their collaboration. Then, a QMIX module attempts to add global state information to $\mathbf{Q}'$ with $\mathbf{Q}'$ as an input.
As shown in Figure~\ref{statemodule}, $Q_{tot}$ represents the joint action value, which can be formulated as:
\begin{equation}
\begin{split}
    \mathbf{Q}''& = \text{elu}(\text{abs}(\text{MLP}(s_t)) \odot \mathbf{Q}' + \text{MLP}(s_t)) \\
    Q_{tot}& = \text{sum}(\text{abs}(\text{MLP}(s_t))\odot \mathbf{Q}'')) + \text{MLP}(s_t)
\end{split}
\label{stateeq}
\end{equation}

\subsection{Loss function}
Our loss function follows the TD-error: %~\cite{sutton2018reinforcement}:
\begin{equation}
    \mathcal{L}(\theta) = \frac{1}{2} (y_{tot} - Q_{tot}(\bm{\tau}, \bm{u}|\theta))^2,
\end{equation}
where the joint action-value function is parameterized by $\theta$, and $y_{tot} = r + \gamma \max_{\bm{u'}}Q_{tot}(\bm{\tau'}, \bm{u'}|\theta^-)$. The target network is parameterized by $\theta$.
\section{Experiments}
In this section, firstly, we introduce settings in our experiments.
Secondly, we conduct experiments on SMAC, which is commonly adopted for MARL algorithms. To demonstrate the validity of HGCN-MIX, we compare it with several well-known baseline algorithms: QMIX, QTRAN, VDN, RODE, and ROMA. Thirdly, we perform HGCN-MIX with a different number of hyperedges to explore the influence of hyperedges. Lastly, we conduct an ablation to explore the effectiveness of the learning part.
% \begin{table*}[htbp]
% \caption{The maximum median win ratios during training. }
% \label{table1}
% \centering
% \begin{tabular}{c|cccccccccccc}
% \hline\\
% \multirow{2}{*}{Algorithms} &
% \multicolumn{12}{c}{Maximum Median Win Ratios(\%)}\\[3pt]
% \cline{2-7}
% &2s3z&{\tt 56m}&{\tt 27m}&{\tt 3M2}&{\tt 1c}&{\tt 2c}&{\tt 89m} & {\tt 3M} &25m&3s5z&{\tt b-g}&{\tt bane}\\[3pt]
% \hline\\
% HGCN-MIX&\textbf{\underline{100}}&\textbf{\underline{81}}&\textbf{\underline{69}}&\textbf{\underline{94}}&97\\
% \hline\\
% QMIX&\textbf{\underline{100}}&69&31&78&\textbf{\underline{100}}\\
% \hline\\
% VDN&\textbf{\underline{100}}&\textbf{\underline{81}}&22&3&94\\
% \hline\\
% QTRAN&94&59&13&0&56\\
% \hline\\
% RODE&\textbf{\underline{100}}&56&38&72&97\\
% \hline\\
% ROMA&94&25&0&19&50\\
% \hline\\
% \end{tabular}
% \begin{tablenotes}
% \item In table \ref{table1}, the bold underlined win ratio denotes the best performance over 6 methods in each scenario. Scenario names are aggregated to save space. For example, 89m refers to $8m\_vs\_9m$ and {\tt 3M2} refers to $MMM2$ and {\tt bane} refers to $bane\_vs\_bane$.
% \end{tablenotes}
% \end{table*}
\begin{table}[!t]
\caption{Part of The maximum median win ratios during training. }
\label{table1}
\centering
\begin{tabular}{c|cccccc}
\hline\\
\multirow{2}{*}{Algorithms} &
\multicolumn{6}{c}{Maximum Median Win Ratios(\%)}\\[3pt ]
\cline{2-7}\\[-2pt]
&2s3z&{\tt 56m}&{\tt 27m}&{\tt 3M2}&{\tt 1c3s5z}&{\tt 2c64zg}\\[2pt] %&8m\_vs\_9m&MMM&25m&3s5z&so\_many\_baneling&bane\_vs\_bane\\
\hline\\[-1pt]
HGCN-MIX&\textbf{\underline{100}}&\textbf{\underline{81}}&\textbf{\underline{69}}&\textbf{\underline{94}}&97&\textbf{\underline{71}}\\
\hline\\
QMIX&\textbf{\underline{100}}&69&31&78&\textbf{\underline{100}}&69\\
\hline\\
VDN&\textbf{\underline{100}}&\textbf{\underline{81}}&22&3&94 & 44\\
\hline\\
QTRAN&94&59&13&0&56&9\\
\hline\\[-1pt]
RODE&\textbf{\underline{100}}&56&38&72&97&38\\
\hline\\
ROMA&94&25&0&19&50&13\\
\hline\\[-2pt]
Algorithms&{\tt 89m}&{\tt 3M}&25m&3s5z&{\tt somany}&{\tt bane}\\
\hline\\
HGCN-MIX&\textbf{\underline{97}}&\textbf{\underline{100}}&\textbf{\underline{100}}&\textbf{\underline{100}}&\textbf{\underline{100}}&\textbf{\underline{100}}\\
\hline\\
QMIX&\textbf{\underline{97}}&\textbf{\underline{100}}&\textbf{\underline{100}}&\textbf{\underline{100}}&\textbf{\underline{100}}&\textbf{\underline{100}}\\
\hline\\
VDN&\textbf{\underline{97}}&\textbf{\underline{100}}&97&91&\textbf{\underline{100}}&94\\
\hline\\
QTRAN&69&91&62&16&\textbf{\underline{100}}&\textbf{\underline{100}}\\
\hline\\
RODE&94&\textbf{\underline{100}}&\textbf{\underline{100}}&84&\textbf{\underline{100}}&\textbf{\underline{100}}\\
\hline\\
ROMA&53&97&0&0&94&81\\
\hline
\end{tabular}
\begin{tablenotes}
\item The bold underlined winning rate in each column denotes the best performance over 6 methods in each scenario. Scenario names are aggregated to save space. For example, {\tt 89m} refers to $8m\_vs\_9m$ and {\tt 3M2} refers to $MMM2$, {\tt bane} refers to $bane\_vs\_bane$, {\tt somany} refers to $so\_many\_baneling$.
\end{tablenotes}
\end{table}
\begin{figure*}[htbp]
\centering
% \vspace{-0.5 cm}
\subfigure[2s3z]{
    \includegraphics[width = 1.8 in]{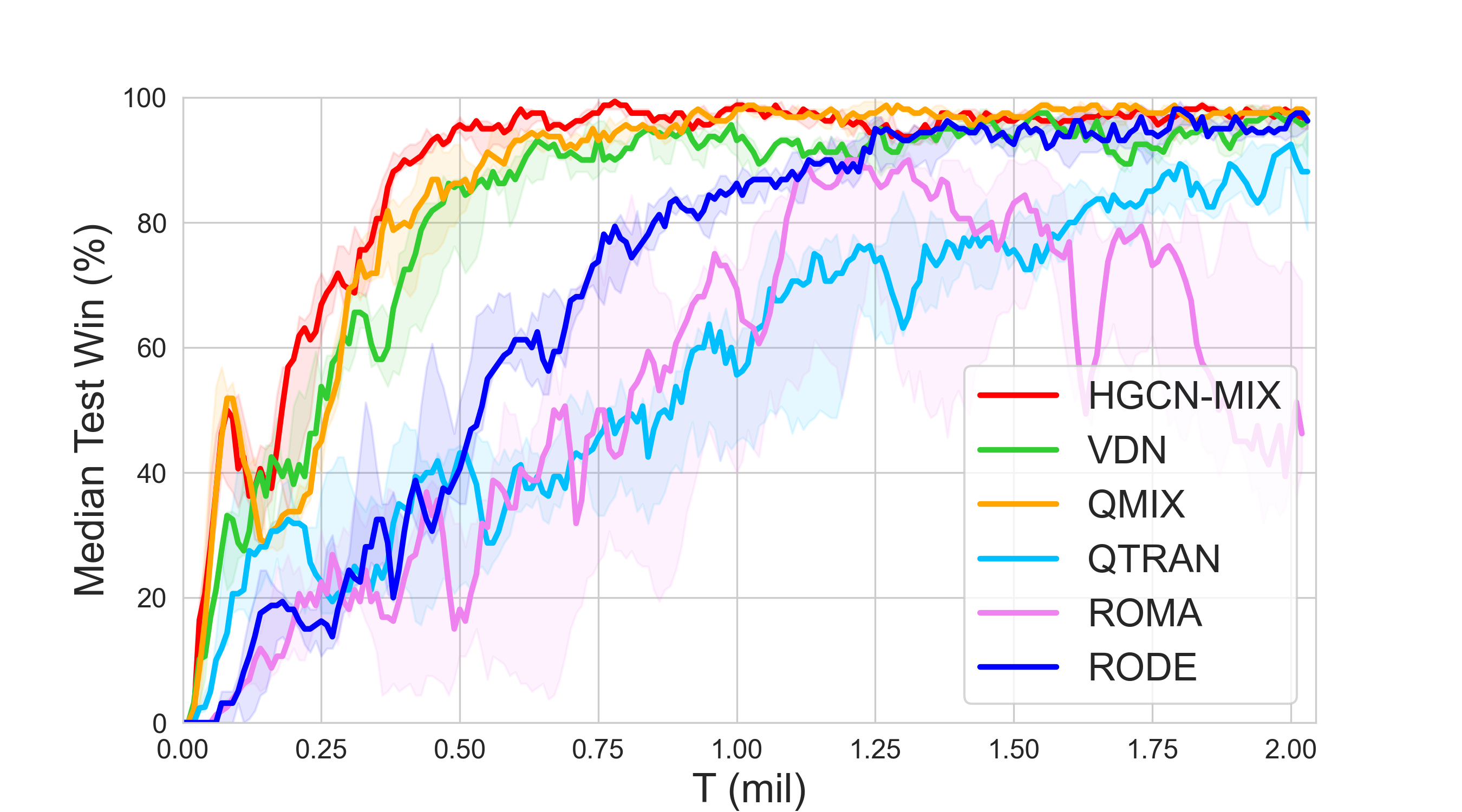}
}
\hspace{-0.6cm}
\subfigure[5m$\_$vs$\_$6m]{
    \includegraphics[width = 1.8 in]{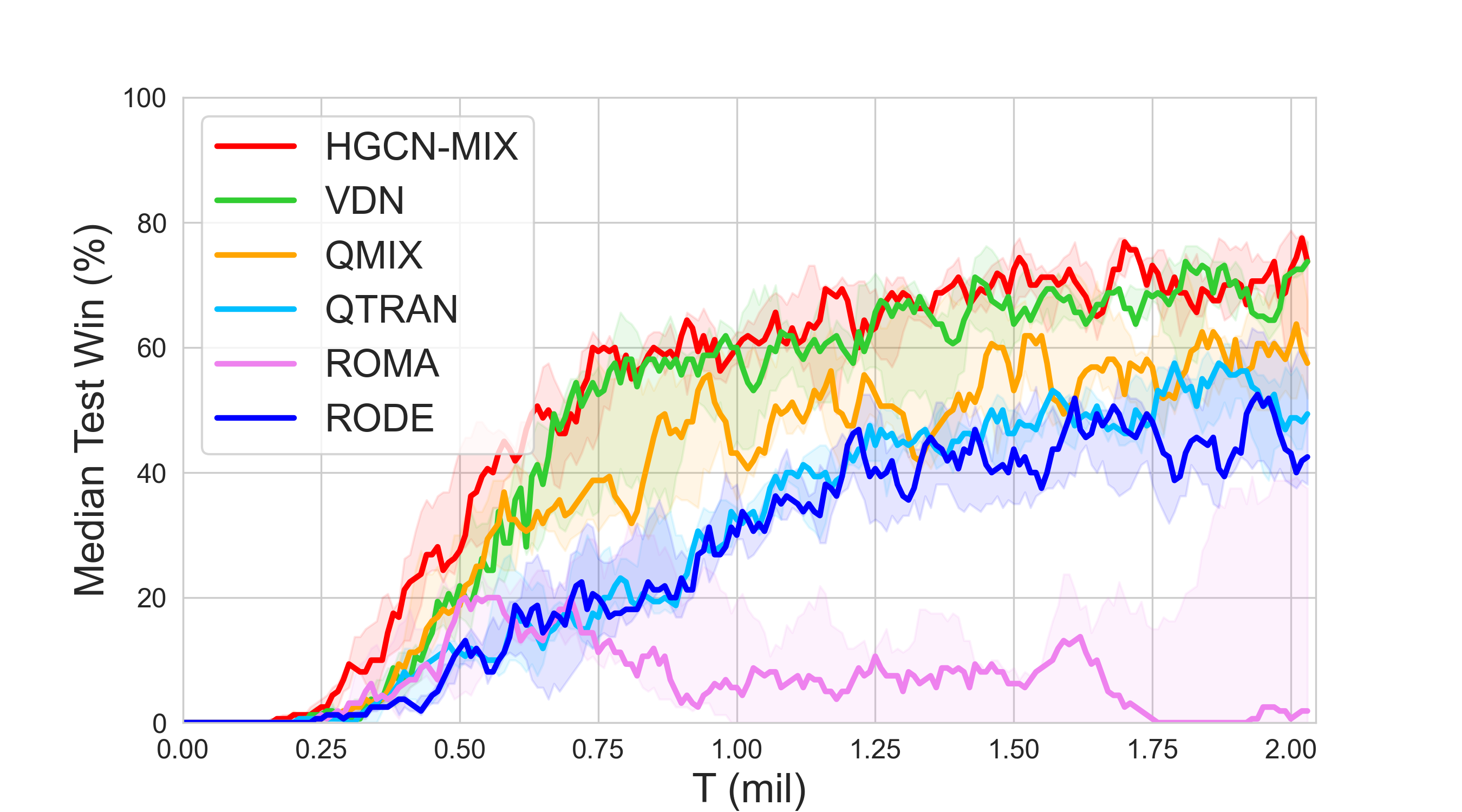}
}
\hspace{-0.6 cm}
\subfigure[27m$\_$vs$\_$30m]{
    \includegraphics[width = 1.8 in]{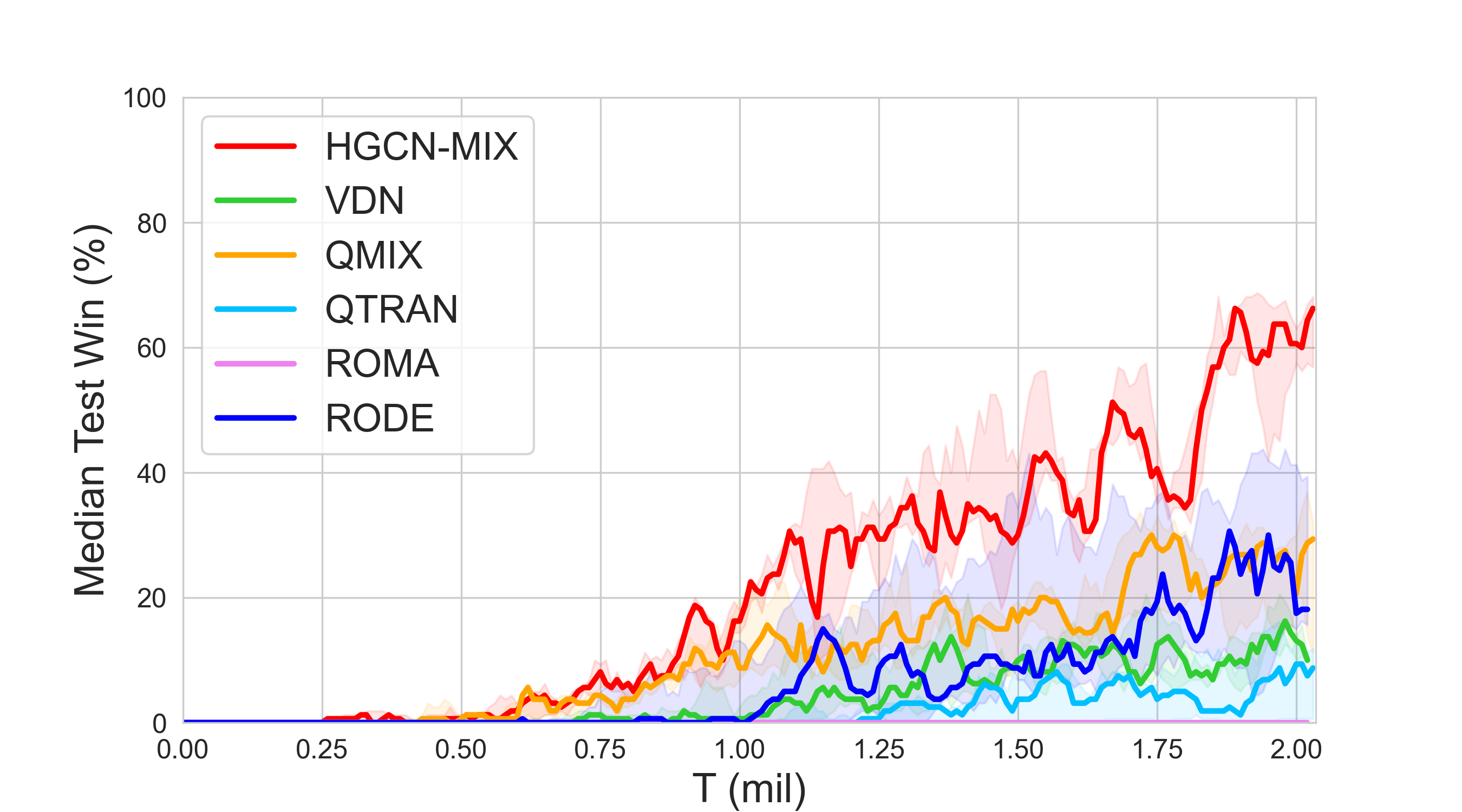}
}
\hspace{-0.6cm}
\subfigure[MMM2]{
    \includegraphics[width = 1.8 in]{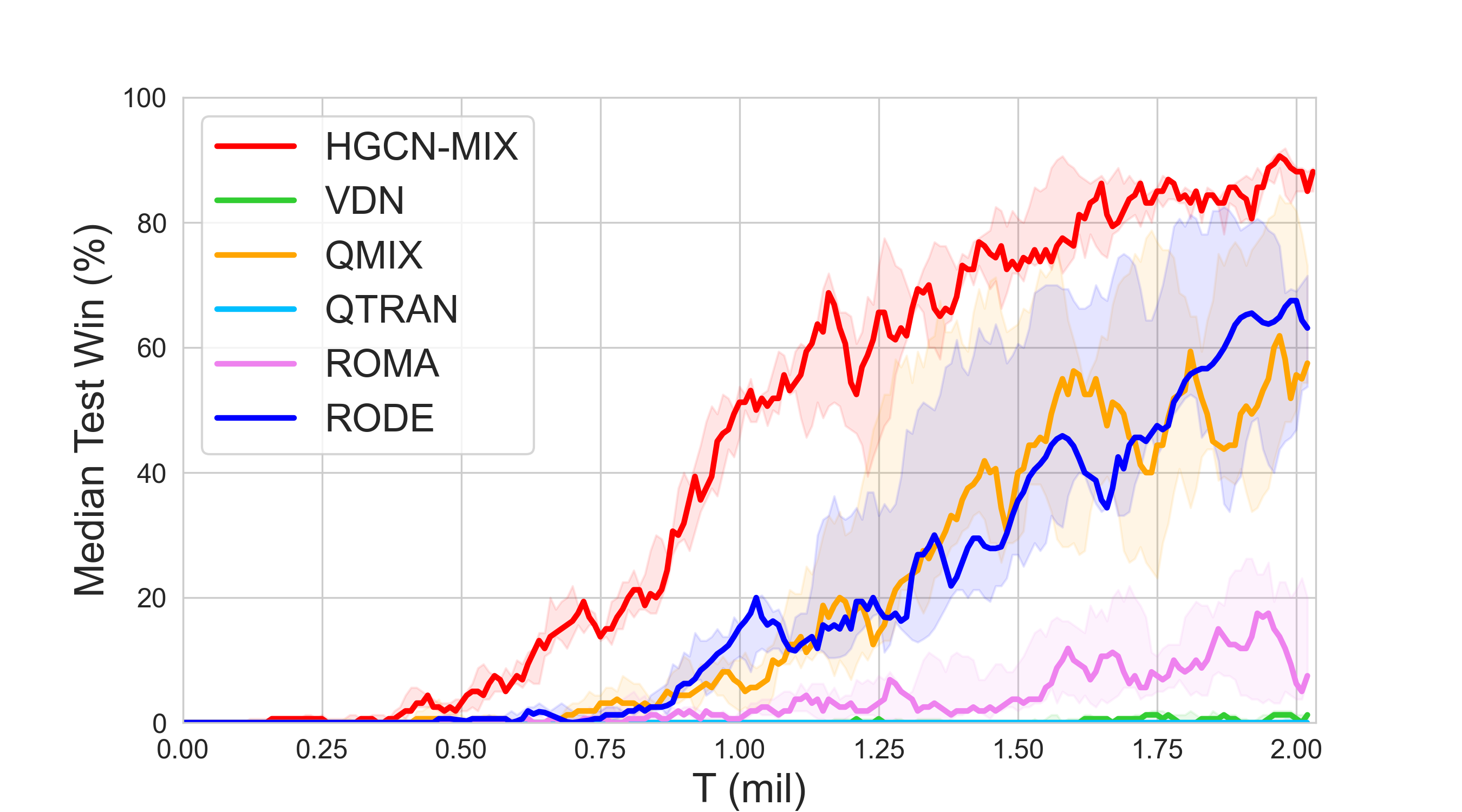}
}
% \vspace{-0.4 cm}

% \vspace{0.5 cm}
\subfigure[1c3s5z]{
    \includegraphics[width = 1.8 in]{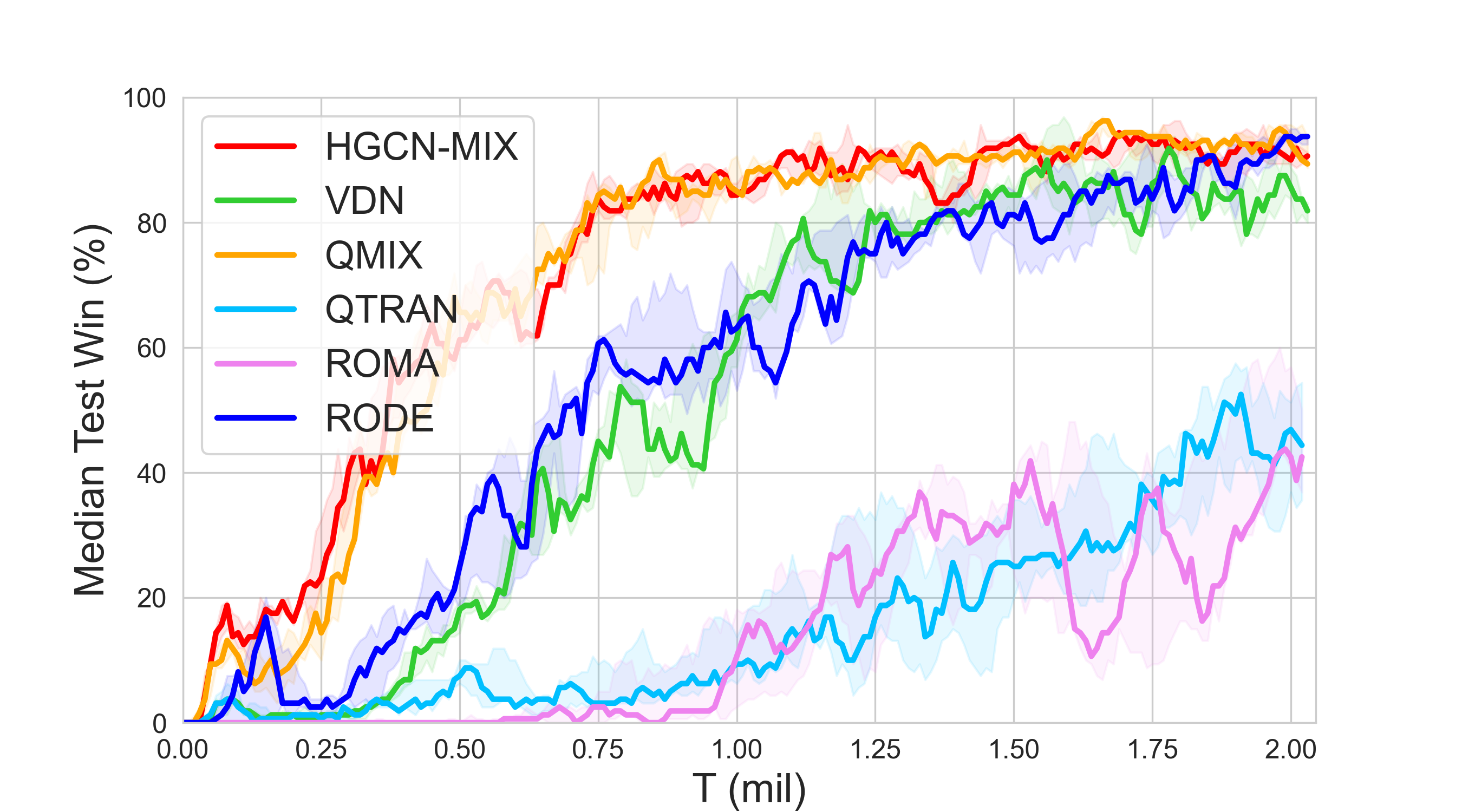}
}
\hspace{-0.6cm}
\subfigure[2c$\_$vs$\_$64zg]{
    \includegraphics[width = 1.8 in]{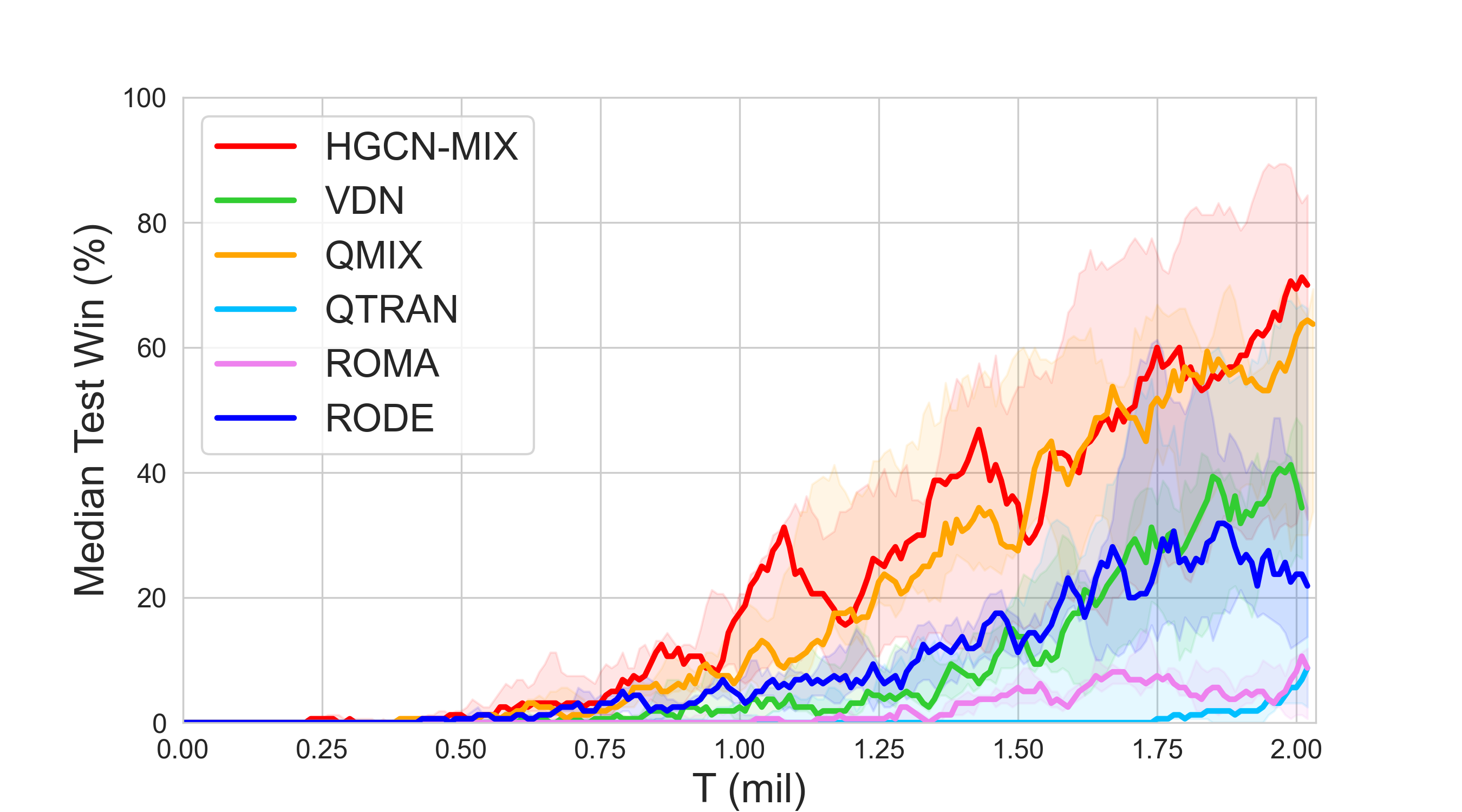}
}
\hspace{-0.6 cm}
\subfigure[8m$\_$vs$\_$9m]{
    \includegraphics[width = 1.8 in]{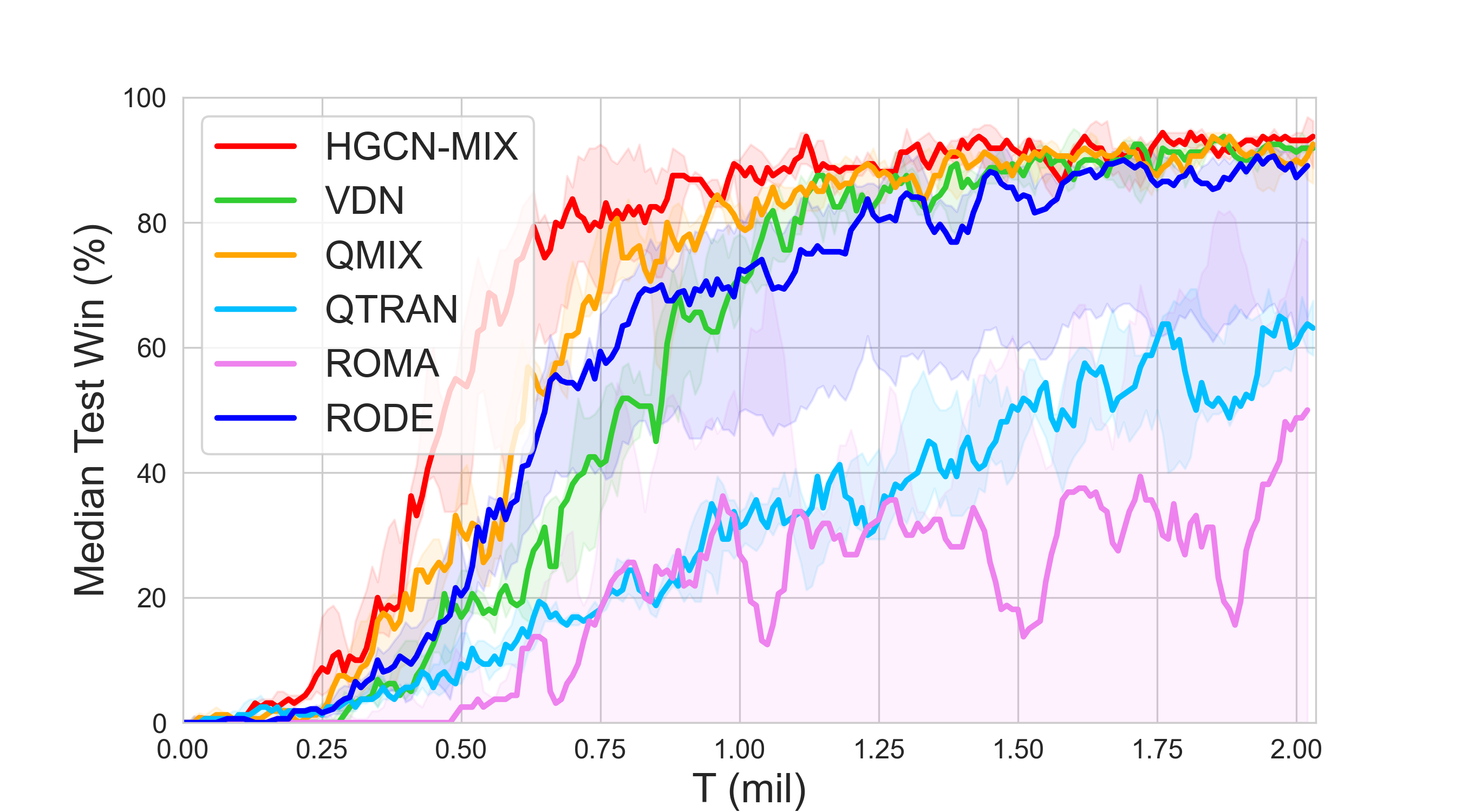}
}
\hspace{-0.6cm}
\subfigure[MMM]{
    \includegraphics[width = 1.8 in]{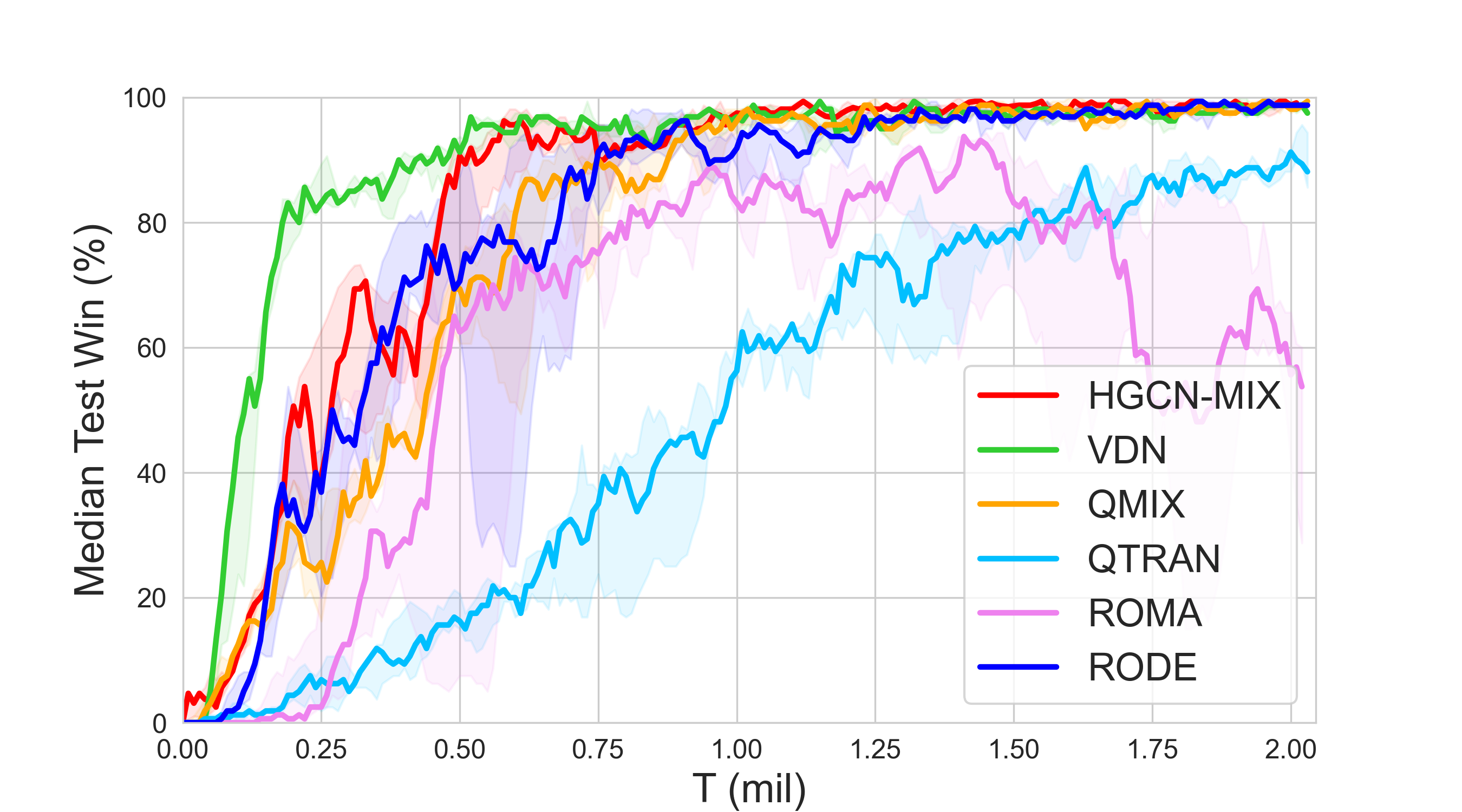}
}

\subfigure[25m]{
    \includegraphics[width = 1.8 in]{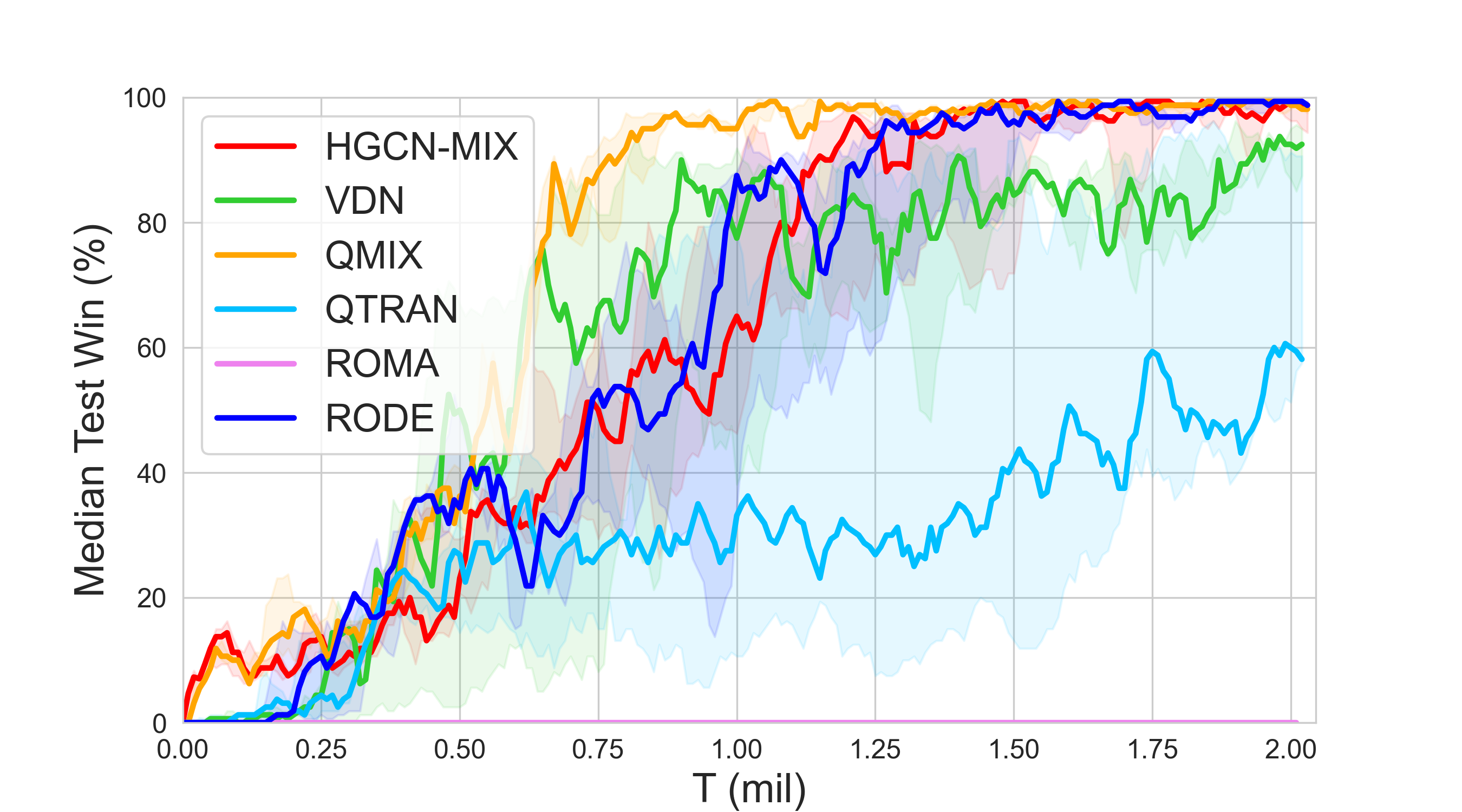}
}
\hspace{-0.6cm}
\subfigure[3s5z]{
    \includegraphics[width = 1.8 in]{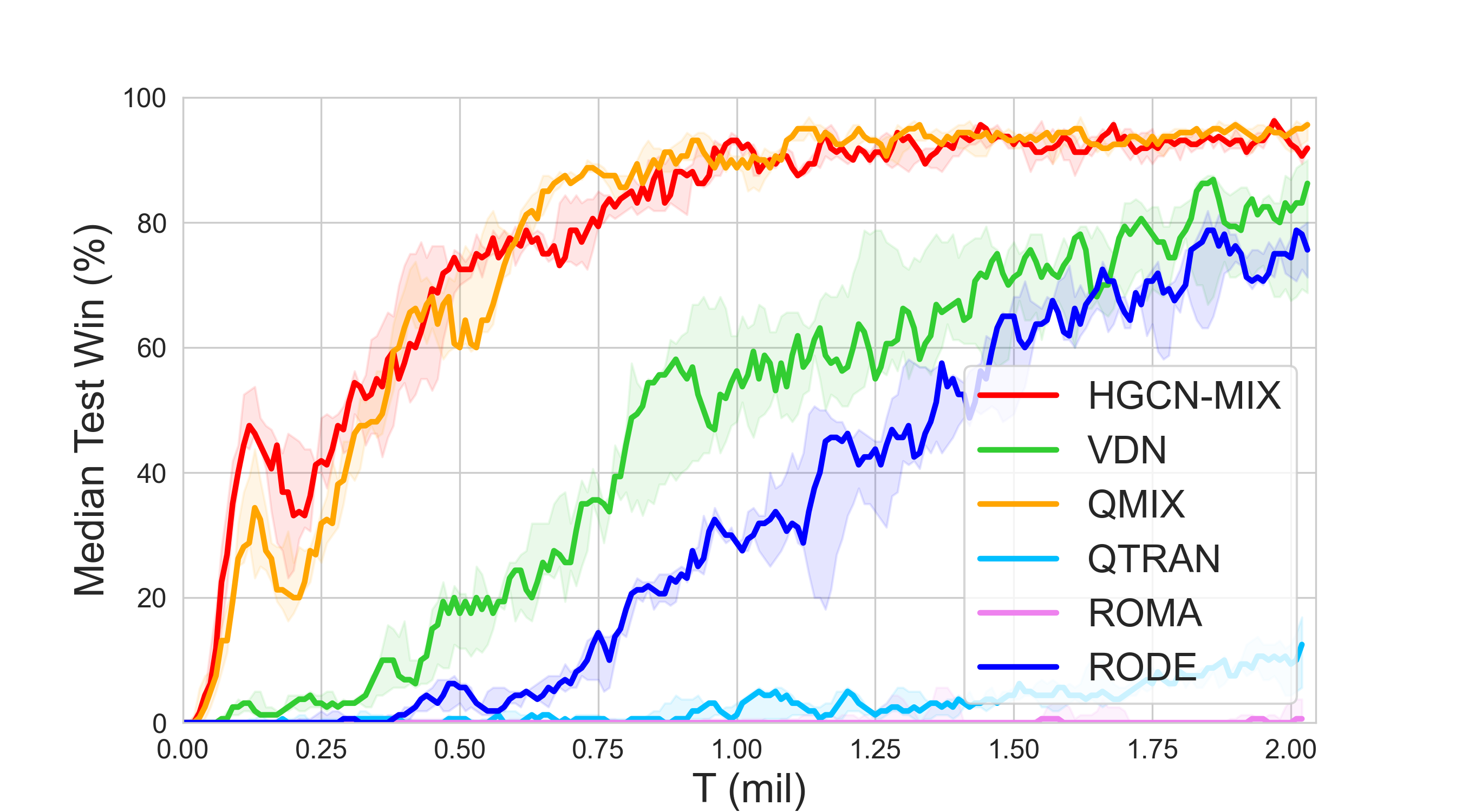}
}
\hspace{-0.6cm}
\subfigure[so\_many\_baneling]{
    \includegraphics[width = 1.8 in]{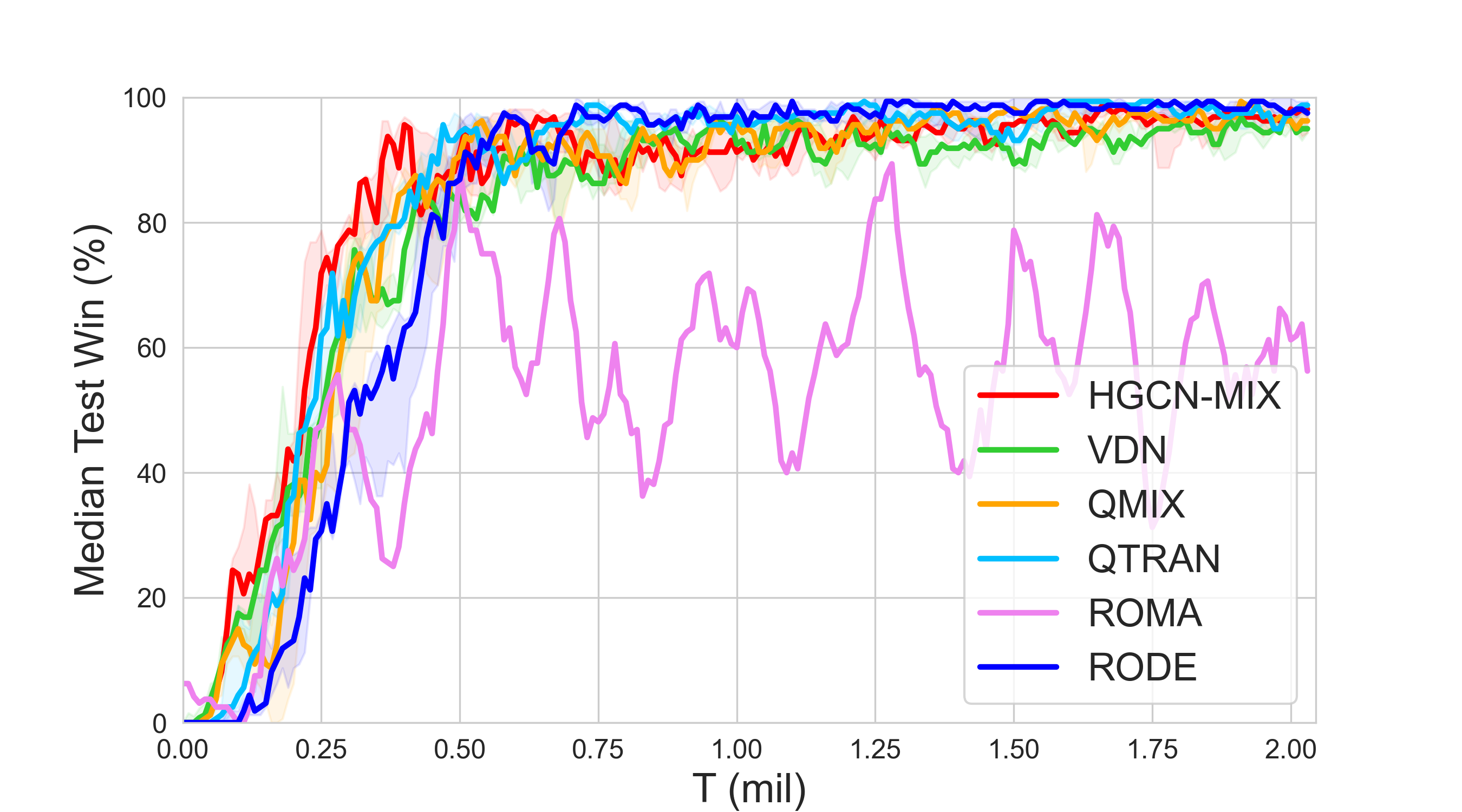}
}
\hspace{-0.6cm}
\subfigure[bane\_vs\_bane]{
    \includegraphics[width = 1.8 in]{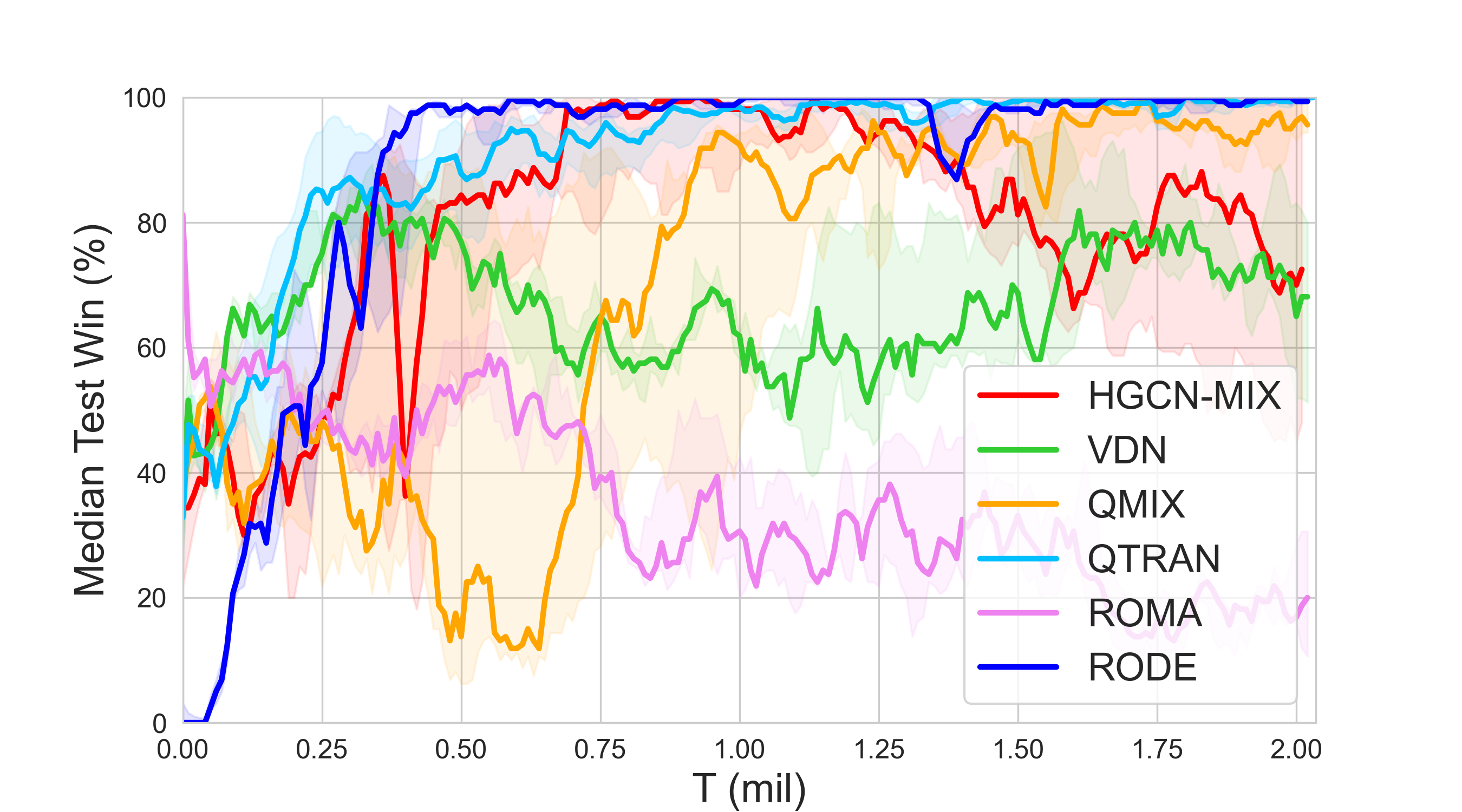}
}
% \vspace{-0.4 cm}
\caption{Overall results in different scenarios. Exepet for $MMM2, MMM, so\_many\_baneling$, and $bane\_vs\_bane$, each digit represents the number of units and each letter corresponds to a type of unit (i.e., 27m means 27 Marines) in scenario name. There are two super hard scenarios: $MMM2, 27m\_vs\_30m$ and three hard scenarios: $2c\_vs\_64zg, bane\_vs\_bane,$ and $5m\_vs\_6m$. The rest scenarios are all easy scenarios. Notice that in those scenarios with special names, $MMM2, MMM, so\_many\_baneling$, and $bane\_vs\_bane$ are all heterogeneous. $MMM2, so\_many\_baneling$ are asymmetric.}
\label{results}
% \vspace{-0.5 cm}
\end{figure*}
\subsection{Settings}
We conduct the experiments on SMAC~\cite{samvelyan2019starcraft} platform, which are commonly adopted to evaluate MARL algorithms.
Our implementation is based on the framework for deep MARL algorithms, i.e., Pymarl~\cite{samvelyan2019starcraft}. Pymarl uses SMAC as the game environment and includes the implementations of QMIX, QTRAN, and VDN. 
The version of the Starcraft II is 4.6.2 (B69232) in our experiments\footnote{Performance is \textbf{NOT} usually comparable between versions~\cite{samvelyan2019starcraft}.}.
We select several well-known baseline algorithms include RODE~\cite{wang2020rode}, ROMA~\cite{wang2020roma}, QMIX, QTRAN~\cite{son2019qtran}, and VDN, as our baseline methods.
% We explain that the reasons why we use StarCraftII 4.6.2 rather than StarCraftII 4.10 as our platform from two aspects. Firstly, scenarios in SC2 4.6.2 are harder than the same scenarios in SC2 4.10 due to the version update. Secondly, the results of baseline methods, including QMIX, Qtran and VDN in SMAC paper\cite{samvelyan2019starcraft} use StarCraftII 4.6.2 not StarCraftII 4.10.
% Therefore, we utilize SC2 4.6.2 as our experimental environments. We also restart the experiments with five different random seeds and record the average winning rates accordingly.
Our experiments are carried out on Nvidia GeForce RTX 3090 graphics cards and Intel (R) Xeon (R) Platinum 8280 CPU. The policies are trained for 2 million timesteps.% The duration of each experiment ranges from 6 to 20 hours, depending on the complexity of the experimental scenarios, e.g., each experiment in $2s3z$ only runs for 6 hours and $27m\_vs\_30m$ runs for 20 hours. 
In HGCN-MIX, the number of learning hyperedge is set to 32. Other hyperparameters and the code of HGCN-MIX follow this link\footnote{https://github.com/cugbbaiyun/HGCN-MIX}. We record the maximum median wining rates during the 2 million timesteps, as we shown in Table~\ref{table1}. 

% \vspace{-0.5 in}
\subsection{Validation}
In Figure~\ref{results}, the solid line is the median winning rates of the 5 random seeds, and the shadow denotes the $25-75\%$ percentiles of the winning rates. The experimental results illustrate that HGCN-MIX outperforms the QMIX approach significantly. In $27m\_vs\_30m$, and $MMM2$, HGCN-MIX achieves the $20\%$ higher winning rate than baseline (i.e., QMIX). In $2s3z, 5m\_vs\_6m$, and $8m\_vs\_9m$, the winning rates trend of HGCN-MIX has plateaued faster than baseline, however, in $25m, 3s5z$, and $MMM$, we observe the opposite results. The learning of hypergraphs in HGCN-MIX leads to extra computational cost.
In the rest 3 out 7 easy scenarios, e.g., $1c3s5z$, $3s5z$, etc., HGCN-MIX reaches a similar performance as baseline methods.

It is noticed that in the hard scenario $2c\_vs\_64zg$, the curve of winning rate in HGCN-MIX convergent unstably. There are two reasons for this phenomenon: 1) The number of ally units is 2, due to the mechanism of the game, and the two cikissis usually take the contrast actions in most scenarios. For example, a cikissi goes down and the other one always goes up. It is difficult for HGCN to learn the coordination between these two cikissis. 2) The count of enemy units is 64. A number of enemy units lead to a large action space, so that we can hardly use the 1-dimensional $Q$ value to represent individual observations and actions.

It can not be ignored that matrix multiplication and matrix inversion in HGCN-MIX bring the extra computational cost compared with QMIX. Considering the significant improvement in winning rates,
we believe that this amount of extra cost is acceptable.
% \subsection{Ablation}
% \subsubsection{Hyperedges}
\subsection{Additional experiments}
\begin{figure}[!t]
\centering
\subfigure[MMM2]{
    \includegraphics[width = 1.75 in]{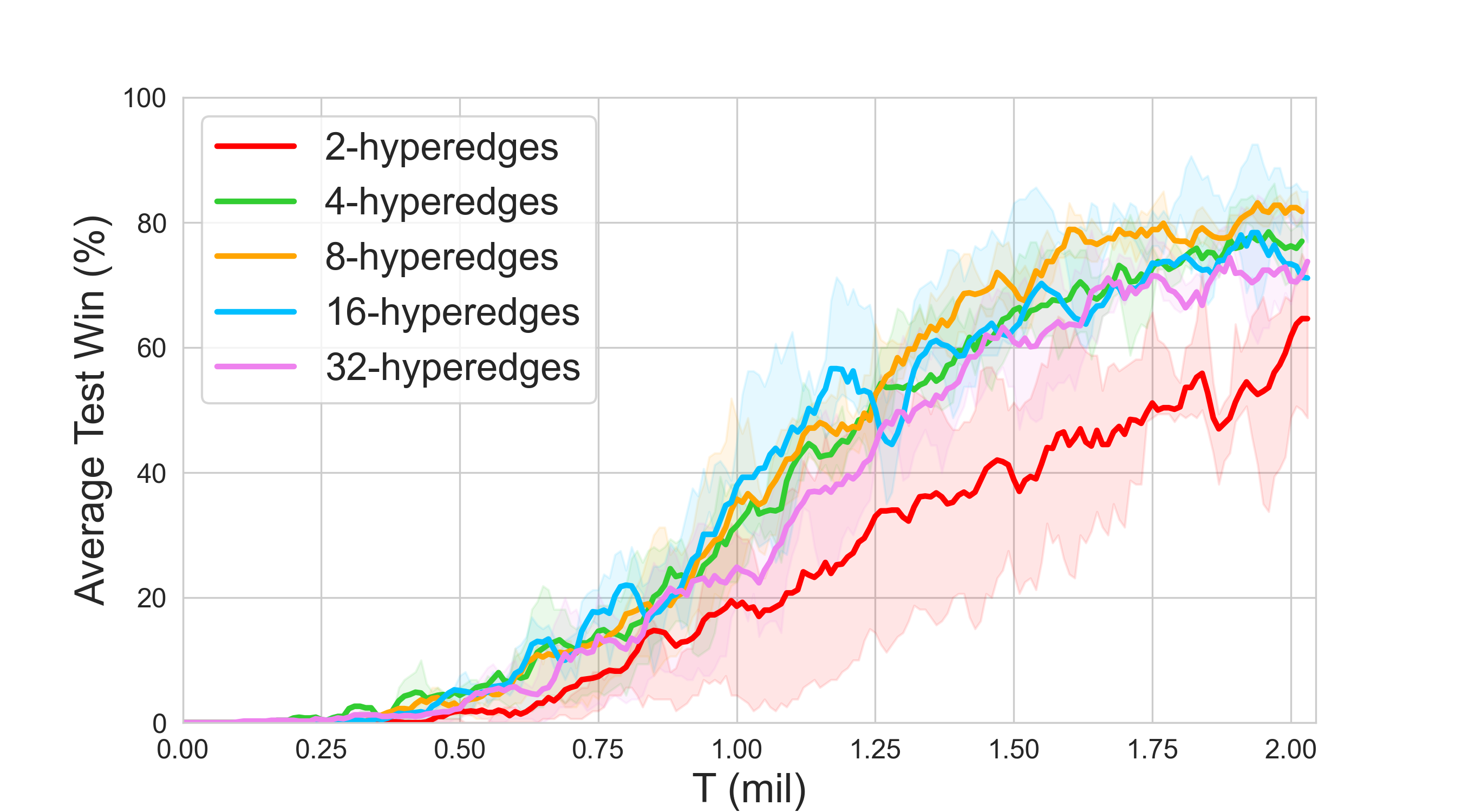}
}
\hspace{-0.8 cm}
\subfigure[27m$\_$vs$\_$30m]{
    \includegraphics[width = 1.75 in]{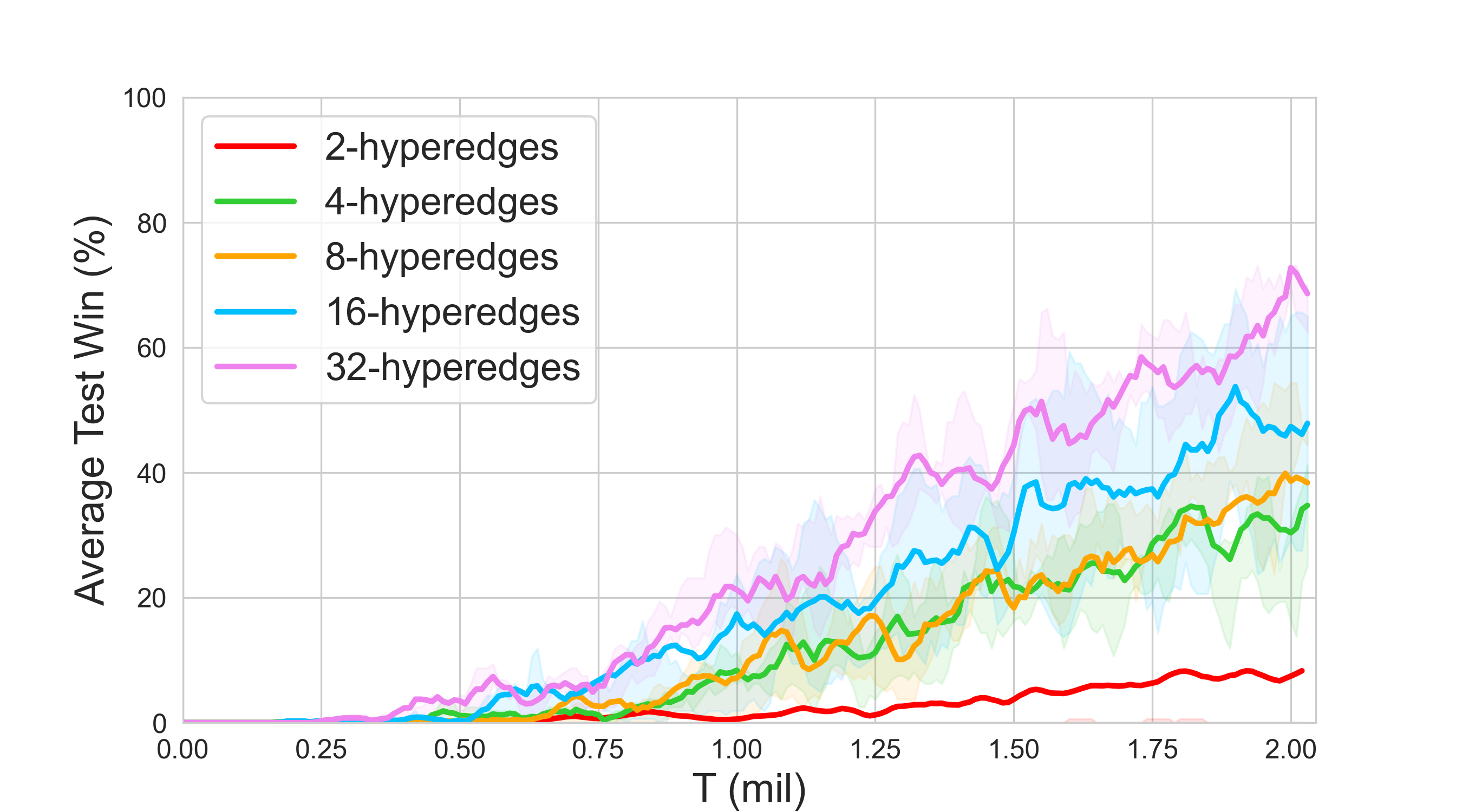}
}
\caption{Average winning rates of agent in two different scenarios. We select the two super hard scenarios: $MMM2$ and $27m\_vs\_30m$. Both $27m\_vs\_30m$ and $MMM2$ are asymmetric and have large counts of agents (10 units in $MMM2$ and 27 units in $27m\_vs\_30m$). Besides, $MMM2$ is heterogeneous but $27m\_vs\_30m$ is homogeneous.}
\label{hyperedgeablation}
\end{figure}
We perform this additional study to examine how the number of hyperedges influencing the performance of joint policies. We set the counts of hyperedges to 2, 4, 8, 16, and 32, respectively. All the hypergraphs learn without one-hot matrices. We test them on the two representative super hard scenarios: $27m\_vs\_30m$ and $MMM2$, and collect the average winning rates as we shown in Figure~\ref{hyperedgeablation}.

\begin{figure}[!t]
\centering
\subfigure[The 13th step in MMM2]{
    \includegraphics[width = 1.75 in]{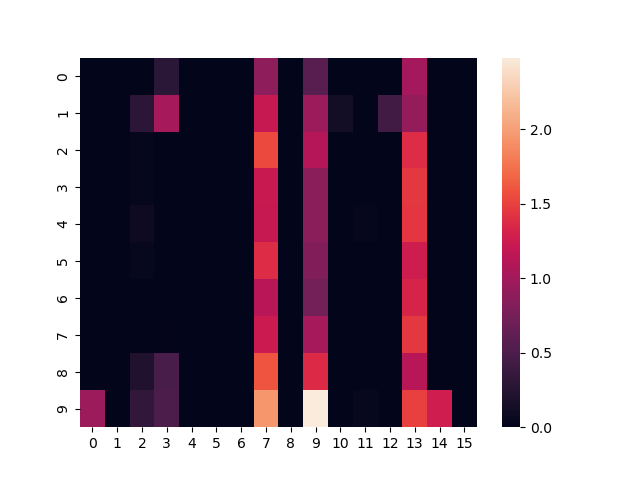}
    \label{13rd}
}
\hspace{-0.8 cm}
\subfigure[The 55th step in MMM2]{
    \includegraphics[width = 1.75 in]{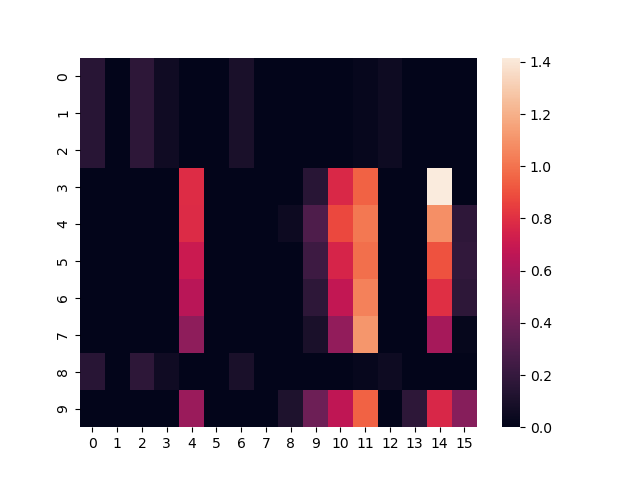}
    \label{55th}
}
\caption{Hypergraphs during training. The two figure show
hypergraphs in different timesteps in the same epsiode. The scenario of the two figures is $MMM2$. Hypergraphs are generated as we have talked about in Section~\ref{buildhyper}. In Figure~\ref{13rd} and~\ref{55th}, line represents agents connected with different hyperedges. Rows show weights in hyperedges, such as row 7 in Figure~\ref{13rd} represent the hyperedge $\epsilon_7$.  }
\label{heatmap}
\vspace{-0.25 in}
\end{figure}

The result in $27m\_vs\_30m$ shows that, with the counts of hyperedges increasing, HGCN-MIX can train a stronger policy. 
Rather than improving the winning rates, the few hyperedges limit the cooperation between different agents. However, in $MMM2$, HGCN-MIX with few hyperedges also works well. With greater than 4 hyperedges, HGCN-MIX has reached the similar performance as baseline methods. To explore the reason that why HGCN-MIX works well in $MMM2$ with few hyperedges, we record two hypergraphs in different timesteps and the same episode in $MMM2$. We visualize the hypergraphs and present in Figure~\ref{heatmap}.

\textit{How to read Figure~\ref{heatmap}.} The same lines, such as 0, 1, 2, and 8 in Figure~\ref{55th}, represent these agents that have been slain, where the health of the agent is set to 0. Individual observations in dead agents are set to the same fixed mask values (i.e., -1). Therefore, the weights generated by neural networks have the same weights. The rows include many zeros introducing the hyperedges equipped with less cooperation. Besides, we can pay attention to the weights in the same hyperedges rather than different hyperedges. 
The diagonal matrix $W$ (defined in Section~\ref{hyperdef}) balances the relative relationships between different hyperedges.

From Figure~\ref{heatmap}, we find that there are only 4 or 5 hyperedges learned different connections in $MMM2$. This result illustrates that only 4 or 5 essential connections exist between agents in $MMM2$. When the number of hyperedges is greater than 4, HGCN-MIX achieves a similar level of performance with the different number of hyperedges. 

% \vspace{-0.1 in}
\subsection{Ablation}
% \subsubsection{Hyperedges}
\begin{figure}[!t]
\centering
\subfigure[2s3z]{
    \includegraphics[width = 1.75 in]{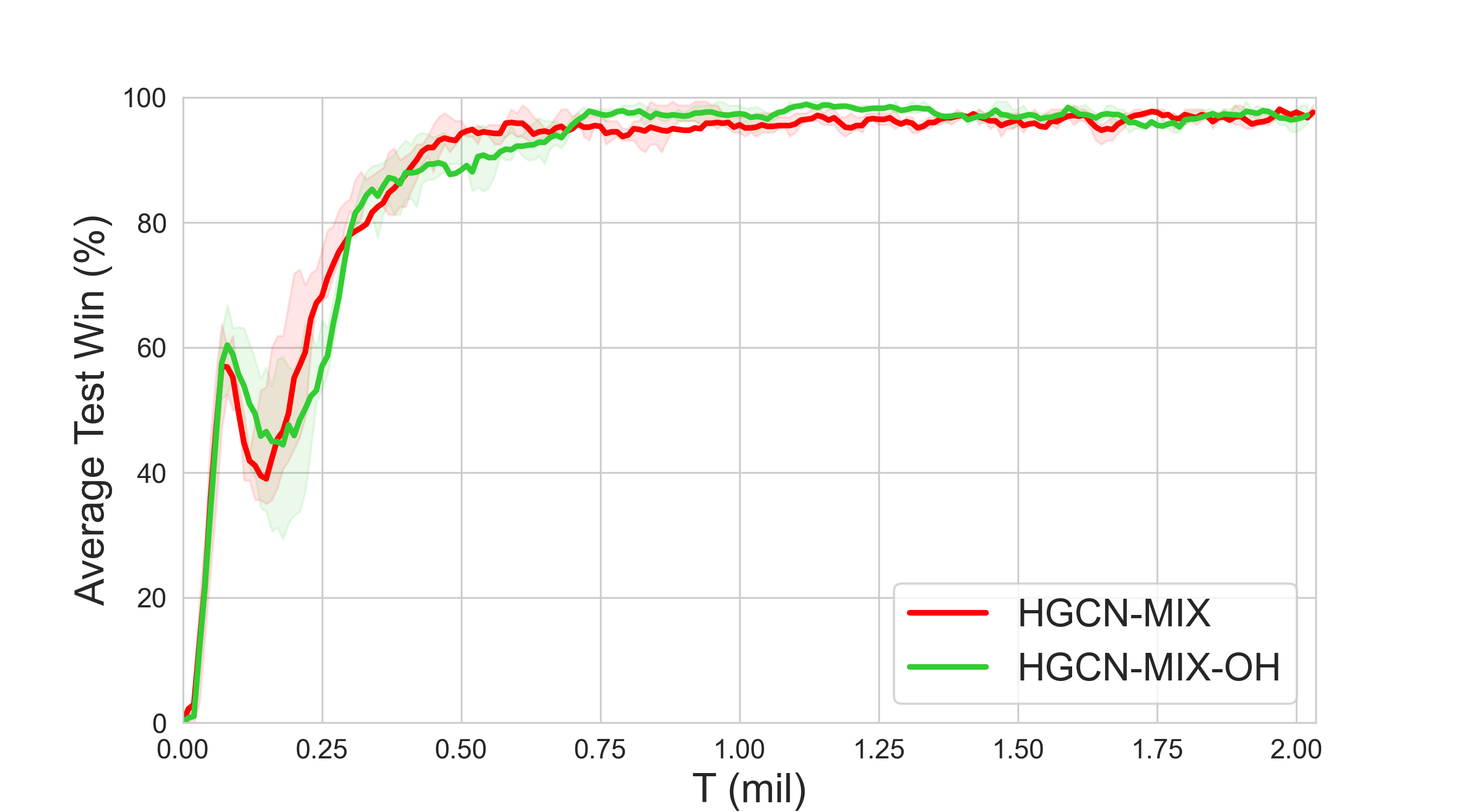}
}
\hspace{-0.8 cm}
\subfigure[MMM2]{
    \includegraphics[width = 1.75 in]{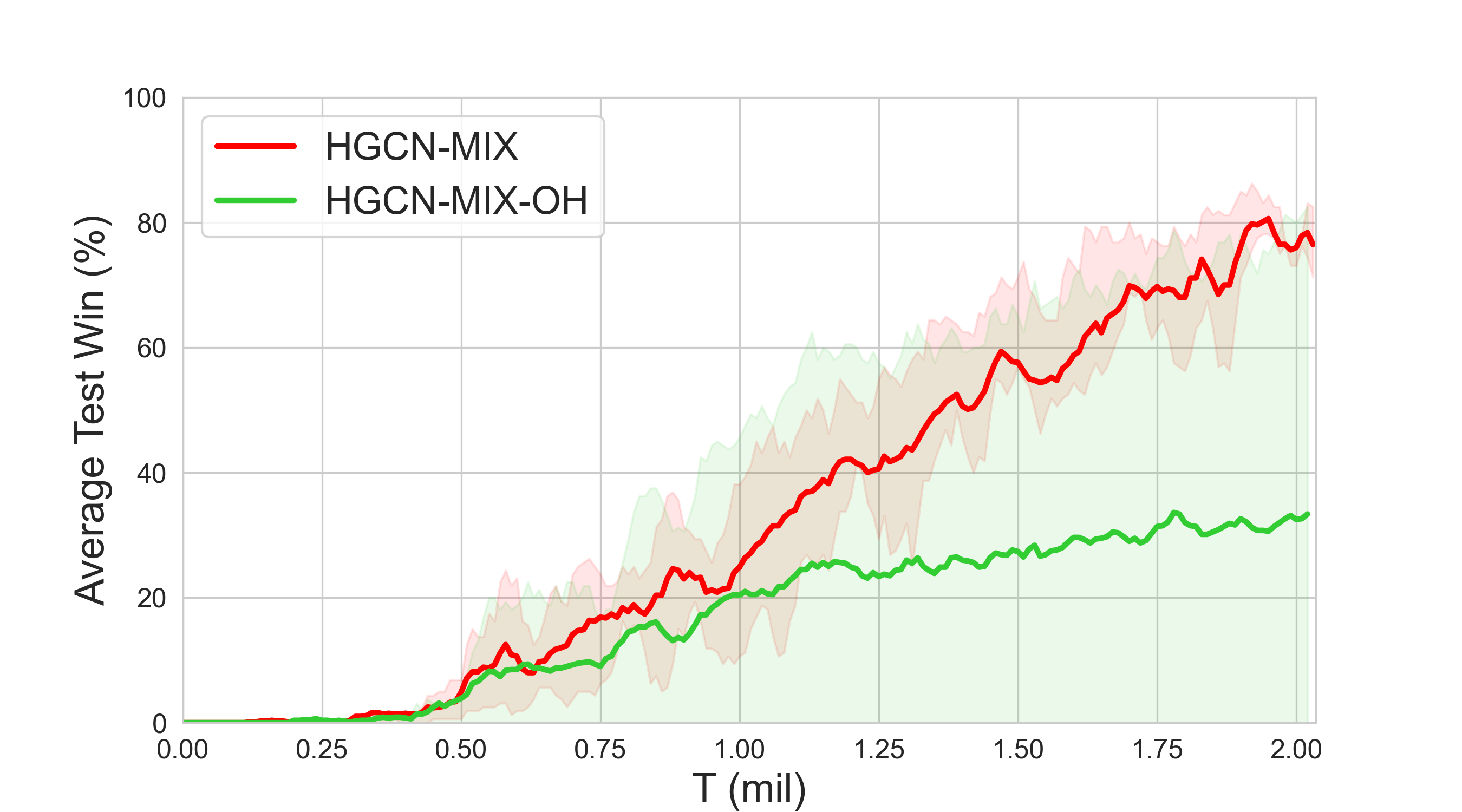}
    \label{learning_part_mmm2}
}
% \vspace{-0.4 cm}
\caption{Average test winning rates of HGCN-MIX and HGCN-MIX-OH in $2s3z$ and $MMM2$.  We select one easy scenario $2s3z$ and one super hard scenario $MMM2$ to explore the influence of the global state module and learning part in the hypergraph. }
\label{ablation}
\vspace{-0.25 in}
\end{figure}

In ablation, we conduct two algorithms (i.e., HGCN-MIX and HGCN-MIX-OH) on $MMM2$ and $2s3z$ to emphasize the influence of the learning part in Eq.~(\ref{HYPERGRAPH}). Hypergraph in HGCN-MIX keeps both one-hot part and learning part, and HGCN-MIX-OH only keeps the one-hot part in Eq.~(\ref{HYPERGRAPH}).
In HGCN-MIX-OH, weights in one-hot part are set to 1. Following Eq.~(\ref{HGCNEQ}), we can derivate that when hypergraph is set as an identity matrix, no matter what $\mathbf{W}$ is, the result of HGCN satisfies: $\bm{x}^{(l+1)} = \bm{x}^{l}$. In this way, HGCN has no effectiveness. In other words, only the global state module works in the mixing network.
We conduct HGCN-MIX and HGCN-MIX-OH on the super hard scenario $MMM2$ and easy scenario $2s3z$, and record the average winning rate of the policy. It is noticed that the hyperparameters of training are the same as that of in HGCN-MIX approach.

As shown in Figure~\ref{ablation}, in $2s3z$, both HGCN-MIX and HGCN-MIX-OH reach almost 100\% winning rates, but in $MMM2$ only HGCN-MIX achieves a high winning rate. HGCN-MIX-OH only works well in some easy scenarios and fails in super hard scenarios. This demonstrates that the global state module is not the most important point for the improvement of performance. The key to enhancing the coordination lies in the learning part.

\section{Conclusions and Future Work}
In this paper, we propose a novel MARL algorithm termed \emph{HyperGraph CoNvolution MIX (HGCN-MIX)}, which combines hypergraph convolution with value decomposition to alleviate the notorious ``lazy agent'' issue. Besides, our approach also overcomes the problem that decomposes joint state-action value functions without any prior knowledge about the connection between nodes (or agents).
By using hypergraph convolution, for each agent, the aggregations of neighborhood information about observations and actions are easily achieved. Based on these information, agents may take better actions to enhance coordination.
We perform HGCN-MIX on SMAC, a well-known benchmark for MARL. Experiments results show that HGCN-MIX outperforms in some scenarios with a large account of agents.
We also conduct ablation experiments to illustrate the connections represented as hyperedges between different agents using neural networks. The learning part in HGCN-MIX plays an important role in HGCN-MIX.
For the sake of HGCN-MIX only transforming the action values, it paves the path for the wide applications in many MARL algorithms based on value decomposition. 

Since HGCN-MIX needs to learn the relationship of coordination between agents, the winning rates trend of HGCN-MIX has plateaued slower than baseline in some scenarios. The problem of speeding up the process of learning the coordination between agents will be studied in future work.
\bibliographystyle{IEEEtran}
\bibliography{refs.bib}
\end{document}